\documentclass[11pt]{article}

\usepackage{graphicx}  
\usepackage{bm}        
\usepackage{pgfplots}
\usepackage{epstopdf}
\usepackage{placeins}
\usepackage{float}
\usepackage{amsthm}
\usepackage{graphicx}  
\usepackage{dcolumn}   
\usepackage{bm}        
\usepackage{amssymb}   
\usepackage{amsmath}
\usepackage{comment}
\usepackage{tikz,ifthen,pgfplots}
\usepackage{color}
\usepackage{hyperref}
\usepackage{placeins}
\usepackage{tabularx}
\usepackage{float}
\usepackage{comment}
\usepackage{algorithmic}

\usepackage{subcaption} 
\usepackage{tikz,ifthen,pgfplots}
\newlength\figureheight 
\newlength\figurewidth 

\usepackage[margin=1in]{geometry}
 
\usepackage[]{amsmath,amssymb,epsfig}
\usepackage{amsthm}
\usepackage{amsmath}
\usepackage{amssymb}
\usepackage{graphicx}
\usepackage{epstopdf}
\usepackage{comment}
\usepackage{array}
\usepackage{algorithm}
\usepackage{url}
\usepackage{bm}

\usepackage{lscape}
\usepackage{setspace}
\usepackage{multicol}
\usepackage{multirow}
\usepackage{color}
\usepackage{colortbl}
\usepackage{xcolor}
\usepackage{mathtools}
\usepackage{booktabs} 

\usepackage{placeins}
\usepackage[%
    font={small,sf},
    labelfont=bf,
    format=hang,    
    format=plain,
    margin=0pt,
    width=0.8\textwidth,
]{caption}
\usepackage[list=true]{subcaption}

\usepackage{enumerate}
\usepackage{tikz}  
\usepackage{tikz-3dplot} 
\usepackage{amssymb}
\usepackage{xifthen}
\usepackage{graphicx}
\usepackage{caption,subcaption}

\usepackage{pgfplots}
\pgfplotsset{compat=newest}
\usetikzlibrary{plotmarks}
\usetikzlibrary{arrows.meta}
\usepgfplotslibrary{patchplots}
\usepackage{grffile}
\usepackage{amsmath}

\newtheorem{theorem}{Theorem}
\newtheorem{proposition}{Proposition}
\newtheorem{lemma}{Lemma}

\DeclareMathOperator{\Tr}{Tr}

\DeclareMathOperator{\rank}{rank}
\DeclareMathOperator{\ddiag}{ddiag}

\DeclareMathOperator{\diag}{diag}
\DeclareMathOperator{\Diag}{Diag}

\newcommand{\calH}{\mathcal{H}}
\newcommand{\rmd}{\mathrm{d}}

\newcommand{\rev}[1]{\textcolor{black}{#1}}

\newcommand*\samethanks[1][\value{footnote}]{\footnotemark[#1]}
\newcommand\myeq{\stackrel{\mathclap{\normalfont\mbox{$\star$}}}{=}}

\title{Positive semi-definite embedding for dimensionality reduction and out-of-sample extensions}

\author{Micha{\"e}l Fanuel \thanks{Université de Lille, CNRS, Centrale Lille, UMR 9189 – CRIStAL, F-59000 Lille, France.}\\
\texttt{michael.fanuel@univ-lille.fr}
\and Antoine Aspeel \thanks{Universit\'e catholique de Louvain, Ecole polytechnique de Louvain,
ICTEAM and CORE, Avenue Georges Lema\^itre, 4-6, Louvain-la-Neuve, B-1348, Belgium.}
\\
\texttt{antoine.aspeel@uclouvain.be}\and Jean-Charles Delvenne\samethanks[2]\\
\texttt{jean-charles.delvenne@uclouvain.be} \and Johan A.K. Suykens\thanks{KU Leuven, Department of Electrical Engineering (ESAT),
STADIUS Center for Dynamical Systems, Signal Processing and Data Analytics,
Kasteelpark Arenberg 10, B-3001 Leuven, Belgium. }\\
\texttt{johan.suykens@esat.kuleuven.be}

}
\date{\today}

\begin{document}
\maketitle

\begin{abstract}
In machine learning or statistics, it is often desirable to reduce the dimensionality of a sample of data points in a high dimensional space $\mathbb{R}^d$. This paper introduces a dimensionality reduction method where the embedding coordinates are the eigenvectors of a positive semi-definite kernel obtained as the solution of an infinite dimensional analogue of a semi-definite program.  This embedding is adaptive and non-linear. We discuss this problem both with weak and strong smoothness assumptions about the learned kernel. A main feature of our approach is the existence of an out-of-sample extension formula of the embedding coordinates in both cases. This extrapolation formula yields an extension of the kernel matrix to a data-dependent Mercer kernel function. Our empirical results indicate that this embedding method is more robust with respect to the influence of outliers, compared with a spectral embedding method.
\end{abstract}

\section{Introduction}

Dimensionality reduction is often an essential step which precedes, for instance, a clustering procedure. This process consists in mapping a sample of $n$ data points in $\mathbb{R}^d$ into a lower dimensional space. 
Among the possible approaches, this work addresses a special case of non-linear \emph{adaptive} embedding, in the spirit of manifold learning. This is in contrast with linear dimensionality reduction methods such as, e.g., the techniques based on compressed sensing which are data-oblivious.
In the spirit of non-linear methods, our approach is closely related to the Diffusion Maps~\cite{DiffusionMaps,Coifman05geometricdiffusions}. A central object of this definition is a diffusion kernel determining a diffusion process on the points of the dataset such that the probability to diffuse from one point to another is large only if the points are in a common neighbourhood.
Given a distribution of data points in $\mathbb{R}^d$, a diffusion embedding is obtained thanks to the spectral decomposition of the diffusion kernel, which is associated to an integral operator, or simply a square matrix in the discrete setting. Then, the $m$-th largest eigenvalues of the diffusion kernel are selected in order to obtain an approximate embedding in $\mathbb{R}^{m}$, while the $m$ embedding coordinates are given by the associated $m$ eigenvectors.
In this work, we discuss a semi-definite program which is closely related to the spectral problem considered in Diffusion Maps. Beyond the embedding of a training dataset, Diffusion Maps allow for the out-of-sample extension of the embedding map so that any forthcoming point can be naturally embedded. This extrapolation is done thanks to a linear formula relying on the Nystr\"om extension (see e.g.,~\cite{GeomHarm,WilliamSeeger}).

The SDP embedding presented in this work shares many properties with a diffusion embedding, although its out-of-sample extension formula is non-linear.
After introducing the notations used throughout this paper, we present the key results in a simplified setting.
\subsection{Notations}
Integral operators will be written by using upper case letters ($A,B,\dots$) while associated integral kernels are denoted by lower case letters ($a,b, \dots$). Let  $(\mathcal{H}, \langle \cdot, \cdot \rangle)$ be a Hilbert space. A linear operator $A:\mathcal{H}\to \mathcal{H}$ is said to be positive semidefinite (\emph{psd}) if $\langle g, Ag\rangle \geq 0$ for all $g\in \mathcal{H}$, and in that case we write $A\succeq 0$. The nuclear norm of an operator on a Hilbert space is written $\|A\|_* = \Tr(\sqrt{A^* A})$. For convenience, we introduce the following space
\[
  \mathcal{S}(\mathcal{H}) =\{ A:\mathcal{H}\to \mathcal{H} \text{ s.t. } A \text{ is self-adjoint and }  \|A\|_* <\infty \}.
  \]
Denote by $\mathcal{C}^m(X)$ the space of $m$-times continuously differentiable functions on $X\subset \mathbb{R}^d$.
The smoothness of a function $d \in \mathcal{C}^m(X)$ is measured thanks to the following semi-norm 
$
|d|_{X,m}=  \max_{|\alpha|=m} \sup_{x\in X} |\partial^\alpha d(x)|
$
where $\partial^\alpha$ is the partial derivative with respect to the multi-index $\alpha$. We will consider the Sobolev space $W_2^s(X)$ with the following inner product 
$  \langle g, g'\rangle = \langle g, g'\rangle_{L^2(X)} + \sum_{|\alpha| = s} \langle \partial^\alpha g, \partial^\alpha g'    \rangle_{L^2(X)}.$
    
Matrices will be denoted by capital bold letters ($\mathbf{A},\mathbf{B},\dots$) while italic bold letters are used for vectors ($\bm{a},\bm{b},\dots$). Then, we will write $\Diag(\bm{a})$ for the diagonal matrix with diagonal entries given by the entries of $\bm{a}$. Similarly, the vector $\diag(\mathbf{A})$ contains the diagonal elements of the matrix $\mathbf{A}$. Let $\ddiag(\mathbf{A})$ be the diagonal matrix constructed from the diagonal of $\mathbf{A}$ and with zero off-diagonal entries. The all ones column vector is denoted by $\bm{1}$. Also, we write the set of integers $\{1,\dots,n\}=[n].$ Finally, we write $a\lesssim b$ if there exists a constant $c>0$ such that $a<c b$.
\subsection{Setting and outline of the main results\label{sec:1}}
In this paper, we consider the domain $X$ to be a bounded set of $\mathbb{R}^d$. For defining the general variational problem, we take $X$ to be the $\ell_1$-ball  $[-c,c]^d$ with $c>0$ since no smoothness assumption is needed in this case, while $X$ is taken to be a $\ell_2$-ball of diameter $2R>0$ when we discuss the kernelized problem.  The space of symmetric and nuclear operators acting on square-integrable functions on $X$, that we denote by $\mathcal{S}(L^2(X))$ or simply $ \mathcal{S}$ when no ambiguity is possible.
Henceforth, we consider \emph{psd} operators in $\mathcal{S}$ because they are associated to symmetric \emph{psd} kernel functions, as we explain shortly below.
Classically, a Hilbert-Schmidt operator $A$ can be represented as a integral operator $Ag(x) =  \int_X a(x,y)g(y)\rmd y$ where $a\in L^2(X\times X)$. A stronger result exists if $A$ is also nuclear, self-adjoint and \emph{psd}.  Namely, a generalized Mercer Theorem by Scovel and Steinwart~\cite{Steinwart2012}, which is recalled as Theorem~\ref{Thm:NuclearDecomposition} in the appendix, states that for all positive semi-definite operators $A\in\mathcal{S}$, there exists an associated integral kernel $a_M(x,y)$, which is defined \emph{pointwisely}, i.e., for all $x,y\in X$. Such a representative $a_M$ is named here a \emph{Mercer} kernel of $A\in \mathcal{S}$, as indicated by the subscript $\cdot_M$. By contrast, a Hilbert-Schmidt operator does not necessarily admit a kernel that is defined pointwisely.

Let us give some motivations for these definitions in the context of manifold learning.
\paragraph{Motivation from diffusion embedding} In~\cite{Coifman05geometricdiffusions,DiffusionMaps,GeomHarm,Nadler} and subsequent works about diffusion geometry, the multi-scale structure of data has been successfully studied thanks to the spectral properties of diffusion kernels. In the context of Diffusion Maps, an operator $\bar{A}\in \mathcal{S}$ is often defined thanks to its symmetric \emph{diffusion kernel}\footnote{There is an alternative definition of Diffusion Maps in~\cite{DiffusionMaps} which involves a non-symmetric kernel. Both operators are related by a conjugation.}
\begin{equation}
\bar{a}(x,y) = \frac{e^{-\|x-y\|_2^2/\sigma^2}}{\sqrt{m(x)m(y)}}-\sqrt{\frac{m(x)}{m_t}}\sqrt{\frac{m(y)}{m_t}},\label{eq:bar_a_0}
\end{equation} 
where the function $m(x) = \int_{X} e^{-\|x-y\|^2/\sigma^2}\rmd y$ is a density and $m_t = \int_{X} m(x)\rmd x$ is a normalization.
Notice that $\bar{A}$ is \emph{psd} for the following reasons.  Since the Gaussian kernel is strictly positive definite, the first term in~\eqref{eq:bar_a_0} is \emph{psd} and, thanks to the normalization, the dominant eigenfunction of this first term is $\psi^{(0)}(x) = \sqrt{m(x)/m_t}$ with eigenvalue $1$.  Therefore, this eigenfunction is also an eigenvector of $\bar{A}$ but with eigenvalue zero, since the second term in~\eqref{eq:bar_a_0} actually subtracts\footnote{Although it is not present in the classical Diffusion Maps definition, the second term in~\eqref{eq:bar_a_0}  only subtracts an uninformative quantity related to the density of data.} a projector on the space generated by $\psi^{(0)}(x)$.

Diffusion Maps  are defined thanks to the spectral decomposition of $\bar{A}$, while the diffusion distance is associated to the $\ell^2$-distance between two points in the diffusion embedding. Let $\{\lambda^{(\ell)}\}_{\ell\geq 1}$ and $\{\psi^{(\ell)}\}_{\ell\geq 1}$ be respectively the eigenvalues sorted in descending order and the associated eigenfunctions of $\bar{A}$. Then, the diffusion embedding $\Psi:X\to \ell^2$ is defined as $\Psi(x) = \big( \lambda^{(\ell)} \psi^{(\ell)}(x)\big)_{\ell\geq 1}$. As a simple consequence, the squared distance between the embedding of $x$ and $y\in X$ is related to a $L^2(X)$ distance as follows
\[
D(x,y)^2= \|\Psi(x)-\Psi(y) \|_{\ell^2}^2 = \| \bar{a}(x,\cdot)-\bar{a}(y,\cdot)\|_{L^2(X)}^2 = \int_X \big(\bar{a}(x,u)-\bar{a}(y,u)\big)^2\rmd u.
\]
It is then common to compute only the eigenfunctions with the largest eigenvalues to yield a low dimensional embedding~\cite{Coifman05geometricdiffusions}.
In particular, the operator with integral kernel $b_\star (x,y) = \psi^{(1)}(x)\psi^{(1)}(y)$  -- associated with the leading eigenfunction  $\psi^{(1)}$ of $\bar{A}$ --  can be obtained as the solution of the following problem:
$\sup_{B\in \mathcal{S}} \Tr(\bar{A} B)$ subject to $ B \succeq 0$ and $\Tr(B) =1$, as it can be seen by introducing a spectral decomposition of $\bar{A}$.

In analogy with this problem, we propose another trace maximization problem with the same diffusion kernel but which involves additional constraints, i.e., the diagonal values of the kernel are constrained rather than the trace. Hence, to bound the diagonal, we introduce a continuous and \emph{strictly positive} function $d(x)$ that enters the variational problem~\eqref{eq:FormalSDP} hereafter. Our analysis is twofold:
\begin{itemize}
  \item in Section~\ref{sec:generalVP}, an analysis is presented with weak smoothness assumptions. A simple discretization scheme is presented.
  \item in Section~\ref{sec:kernelized_intro}, a kernelized approach is given within a Reproducing Kernel Hilbert Space (RKHS) of continuously differentiable functions. This setting yields stronger statistical guarantees obtained thanks to results from~\cite{Rudi2020global} to which we refer extensively.
\end{itemize}
In particular, the latter approach allows for a built-in out-of-sample extension, whereas the former setting has a less natural extrapolation formula.

\subsubsection{General variational problem\label{sec:generalVP}}
In view of this spectral problem, we define the following maximization problem:
\begin{equation}
\sup_{B\in \mathcal{S}(L^2(X))} \Tr(\bar{A} B) \text{ subject to } B \succeq 0 \text{ and } b_M(x,x) \leq d(x)\quad \text{ almost everywhere},\tag{$\text{VP}$}\label{eq:FormalSDP}
\end{equation}
where $b_M(x,y)$ is a Mercer kernel associated to $B$ as it is given by Theorem~\ref{Thm:NuclearDecomposition} in appendix. The inequality constraint in~\eqref{eq:FormalSDP} should be satisfied on $X$ except possibly on a negligible set for the Lebesgue measure. As a main result of our paper, we show that 
the diagonal constraint makes sense and that the supremum hereabove is attained by a positive semi-definite operator $B_\star\in\mathcal{S}$ with a Mercer kernel satisfying $b_{\star,M}(x,x) \leq d(x)$ almost everywhere.\\
\paragraph{Discrete problem} In practice, we solve an analogue of~\eqref{eq:FormalSDP} involving square symmetric matrices, in order to calculate an embedding given by the eigenvectors of the solution. To start with, we define a discrete analogue of~\eqref{eq:bar_a_0}. Given a sample of data points $\{x_i\in X\}_{i\in [n]}$ and a kernel matrix $[\mathbf{K}]_{ij} = \exp(-\|x_i-x_j\|_2^2 /\sigma^2)$ for $i,j\in [n]$,
we can define the empirical normalized kernel by
\begin{equation}
\mathbf{A} = \Diag(1/\sqrt{\mathbf{m}})\mathbf{K} \Diag(1/\sqrt{\mathbf{m}}) \quad\text{ with } \quad\mathbf{m} = \mathbf{K}\mathbf{1},\label{eq:DiscreteA}
\end{equation}
where the above division is done elementwise.
The subtracted kernel matrix reads
$\mathbf{\bar{A}}= \mathbf{A} - \bm{v}^{(0)} \bm{v}^{(0)\top}$ where $\bm{v}^{(0)} = \sqrt{\mathbf{m}/(\mathbf{1}^\top \bm{m})}$  is the dominant eigenvector of $\mathbf{A}$ with eigenvalue $1$.
Given the function $d(x)$, a discrete counterpart of~\eqref{eq:FormalSDP} is the Semi-Definite Program
\begin{equation}
\max_{\mathbf{B}\succeq 0} \Tr\left(\mathbf{\bar{A}} \mathbf{B}\right), \text{ subject to } \diag(\mathbf{B}) \leq \bm{d} \tag{$\text{SDP}$},\label{eq:SDP}
\end{equation}
where the inequality constraint holds elementwise and with $[\bm{d}]_i = d(x_i)$ for all $i\in [n]$.
The embedding coordinates
$ x_i\mapsto \bm{\Xi}_{i\ast} = [\bm{\chi}^{(1)}_i, \dots ,\bm{\chi}^{(r)}_i ]^\top$ are given by the eigenvectors $\{\bm{\chi}^{(\ell)}\in \mathbb{R}^{n}\}_{1\leq\ell\leq r}$ of the solution $\bm{B}_\star$ with non-zero eigenvalue $\lambda^{(\ell)}$, that are normalized such that $\|\bm{\chi}^{(\ell)}\|^2_{2} = \lambda^{(\ell)}$. The embedding obtained is illustrated in Figure~\ref{Fig:OutliersBandwidth} in a toy example. An advantage of the SDP embedding over the diffusion embedding is that the length of the embedding vectors is constrained by construction. Indeed, the squared length of the embedding vector is the corresponding diagonal element of the matrix $\bm{B}_\star$ that is upper bounded in~\eqref{eq:SDP}. This constraint impedes a localization effect that can be observed in spectral embedding methods (see Figure~\ref{Fig:Wine} for instance).
The dimensionality of the SDP embedding is the rank of $\bm{B}_\star$ which is presumably low as we argue here in the light of our empirical simulations and of Proposition~\ref{Prop:Rank} which is given in Section~\ref{sec:dim_red}.\\
\begin{figure}[t]
\centering
\setlength\figureheight{0.2\textwidth} 
\setlength\figurewidth{0.22\textwidth}
\begin{subfigure}[t]{0.3\textwidth}
\input{Figures/DataOutliers.tikz}
\end{subfigure}
\hfill
\begin{subfigure}[t]{0.3\textwidth}
\input{Figures/EmbeddingDM1p2Outliers.tikz}
\end{subfigure}
\hfill
\begin{subfigure}[t]{0.3\textwidth}
\input{Figures/EmbeddingSDP1p2Outliers.tikz}
\end{subfigure}
\caption{SDP and Diffusion Maps embedding ($\sigma = 1.2$). The dataset -- on the lhs -- contains three clusters with 100 points each and 10 outliers (red stars). \label{Fig:OutliersBandwidth}}
\end{figure}
\paragraph{Out-of-sample extension}
In this work, we also propose an extrapolation of the vectors $\bm{\chi}^{(\ell)} $ which allows to embed additional data arising after the embedding of the initial $n$ datapoints\footnote{The out-of-sample extension problem could also be addressed for~\eqref{eq:FormalSDP}. We leave this question for future work.}.
Firstly, we can define an empirical diffusion kernel function whose Gram matrix $\bm{A}$ is given by~\eqref{eq:DiscreteA}, i.e.,
\begin{equation}
a_e(x,y) = \frac{e^{-\|x-y\|_2^2/\sigma^2}}{\sqrt{m_e(x) m_e(y)}} \text{ and } m_e(x) = \sum_{i=1}^n e^{-\|x-x_i\|_2^2/\sigma^2},\label{eq:EmpiricalNormalizedKernel}
\end{equation}
where we indicated by a subscript $\cdot_e$ an extended empirical quantity.
In the same way, the extension of any column of the Gram matrix of $a_e(x,y)$ is naturally defined by  $[\bm{a}_e(x)]_i = a_e(x,x_i)$ in view of~\eqref{eq:EmpiricalNormalizedKernel}. This yields an extrapolation formula for the subtracted kernel. First, we extend the matrix $\bar{\mathbf{A}} = \mathbf{A} - \bm{v}^{(1)} \bm{v}^{(1)\top}$ by adding one additional row and column as follows,
\begin{equation}
\mathbf{\tilde{A}} = 
\left[\begin{array}{c c}
 \bar{\mathbf{A}} & \bar{\bm{a}}_e(x)\\
\bar{\bm{a}}_e (x)^\top & \alpha(x)
\end{array}\right]\quad \text{ with }\quad \bar{\bm{a}}_e(x) = (\mathbb{I} - \bm{v}^{(1)} \bm{v}^{(1)\top})\bm{a}_e(x),\label{eq:abar}
\end{equation}
and where the diagonal extension is $\alpha(x) = 1/m_e(x) - m_e(x)/(\mathbf{1}^\top \bm{m})$ with $x\in X$ in the light of~\eqref{eq:bar_a_0}.
Then, this extended empirical kernel~\eqref{eq:abar} can serve to define the extension of the SDP embedding. Indeed, the proposed interpolation formula is the \emph{normalized} Nystr\"om extension~\cite{WilliamSeeger},
\begin{equation}
\chi^{(\ell)}_e(x) = \sqrt{d(x)}\frac{\bm{\bar{\bm{a}}}_e^\top (x) \bm{\chi}^{(\ell)}}{\sqrt{\bm{\bar{\bm{a}}}_e^\top(x) \bm{B}_\star \bm{\bar{\bm{a}}}_e(x)} } \text{ for all } x\in \mathcal{D}_X,\label{eq:Out-of-sample2}
\end{equation}
where $\mathcal{D}_X = \{x\in X |  \bm{B}_\star \bm{\bar{\bm{a}}}_e(x)\neq 0\}$. While the numerator in~\eqref{eq:Out-of-sample2} is similar to the Nystr\"om extension, the denominator is here rather different since it can be interpreted as a spherical normalization so that $\sum_{\ell=1}^r(\chi^{(\ell)}_e(x))^2 = d(x)$. In a word, this particular form of the out-of-sample extension is obtained thanks to the optimality conditions of~\eqref{eq:SDP}.  Our second main result, Theorem~\ref{Thm:OOS} given in Section~\ref{sec:oos}, indicates that the interpolation~\eqref{eq:Out-of-sample2} is defined \emph{almost everywhere} on $X$ hereafter, that is, its domain is $\mathcal{D}_X = X\setminus X_0$ where $X_0$ is negligible. In essence, this is mainly due to the properties of the Gaussian kernel in~\eqref{eq:EmpiricalNormalizedKernel}, which belongs to a Hilbert space of analytic functions. Explicitly, the extended embedding reads then
\[
x\mapsto \bm{\Xi}_e(x) = [{\chi}^{(1)}_e(x) , \dots ,{\chi}^{(r)}_e(x) ]^\top.
\]
As a consequence, the extension of the matrix $\mathbf{B}_\star$ to a kernel function can be defined by
$
b_{\star,e}(x,y) = \bm{\Xi}_e(x)\bm{\Xi}_e(y)^\top
$
for all $(x,y)\in \mathcal{D}_X\times \mathcal{D}_X$. The latter expression defines a data-dependent kernel function which has been estimated from the sample of $n$ points in an unsupervised way.
In this paper, we choose the upper bound on the diagonal as $d(x)= \alpha(x)$ 
 in~\eqref{eq:abar}; this choice is motivated by Lemma~\ref{Lemma:diagpos} in Section~\ref{sec:oos}.

\begin{figure}[h!]
\centering
\begin{subfigure}[t]{0.49\textwidth}
\centering
$\bar{a}(x,y)$ \par\medskip
\includegraphics[trim={2cm 8.5cm 2cm 8.5cm},clip,scale = 0.45]{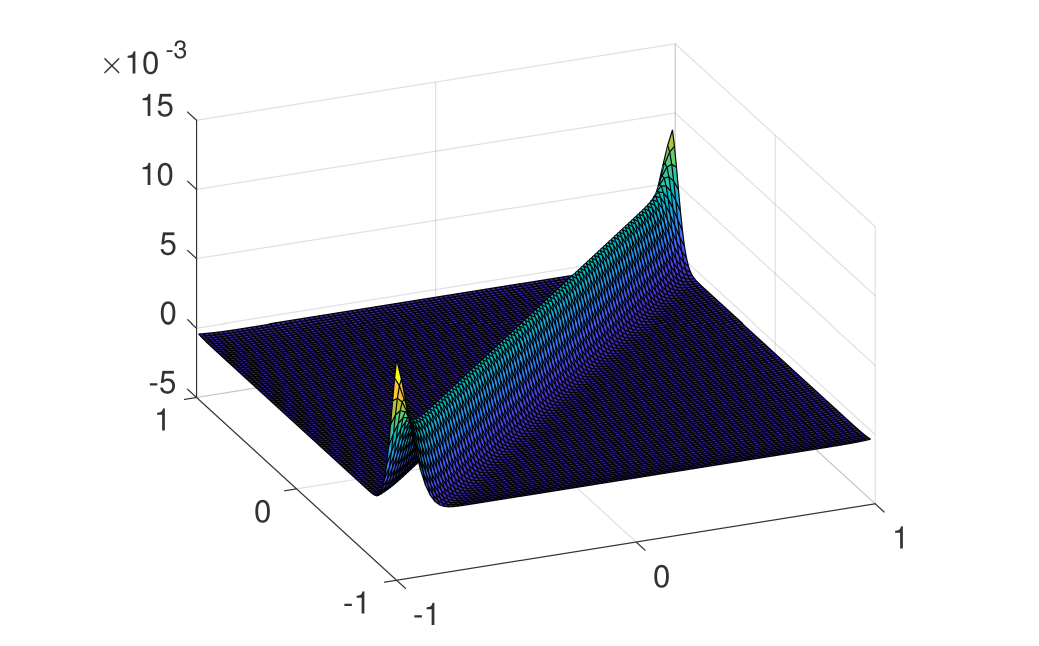}
\end{subfigure}
\begin{subfigure}[t]{0.49\textwidth}
\centering
$b_\star(x,y)$\par\medskip
\includegraphics[trim={2cm 8.5cm 2cm 8.5cm},clip,scale = 0.45]{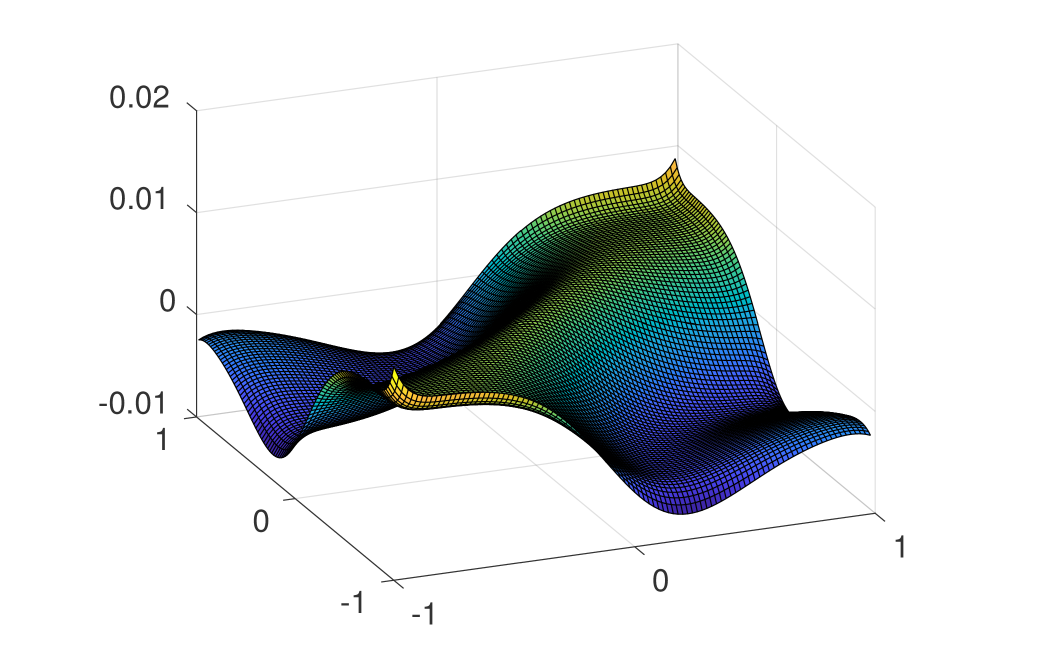}
\end{subfigure}
\caption{For $\sigma = 0.1$, on the left, $\bar{a}(x,y)$ and on the right $b_\star(x,y)$ are plotted on the square $[-1,1]^2$. This plot illustrates the shape of the numerical solution of~\eqref{eq:FormalSDP}.\label{Fig:Toy1}}
\end{figure}

In Figure~\ref{Fig:OutliersBandwidth}, the effect of outliers on Diffusion Maps and SDP embedding is illustrated on an artificial example, where we observe that the outliers remain far from the denser clusters for the SDP embedding, while their positions are not fixed in the Diffusion Maps embedding. Other illustrations of dimensionality reduction are given in Section~\ref{sec:6}.
Furthermore, although this is not an example of dimensionality reduction, the solution of a discrete approximation of~\eqref{eq:FormalSDP} for a simple toy model is also illustrated in Figure~\ref{Fig:Toy1} and Figure~\ref{Fig:Toy2}, where $X = [-1,1]$. Then, the square $[-1,1]^2$ is discretized in a square grid of $2001^2$ points and the result of the discrete optimization problem is displayed in Figure~\ref{Fig:Toy1} for $\sigma = 0.1$. Figure~\ref{Fig:Toy2} illustrates the out-of-sample extension.
\begin{figure}[h!]
\centering
\setlength\figureheight{0.1\textwidth}
\setlength\figurewidth{0.3\textwidth}
\begin{minipage}{0.49\textwidth}
%
%
\definecolor{mycolor1}{rgb}{0.85000,0.32500,0.09800}%
\begin{tikzpicture}

\begin{axis}[%
width=\figurewidth,
height=\figureheight,
at={(0.772in,0.501in)},
scale only axis,
xmin=-3,
xmax=3,
xlabel style={font=\color{white!15!black}},
xlabel={$\chi_1(x)$},
ymin=-2.5,
ymax=2.5,
ylabel style={font=\color{white!15!black}},
ylabel={$\chi_2(x)$},
ylabel near ticks, ylabel shift={-5pt},
xlabel near ticks, xlabel shift={-2pt},
axis background/.style={fill=white},
title style={font=\bfseries},
title={SDP embedding},
axis x line*=bottom,
axis y line*=left,
tick label style={font=\tiny},
]
\addplot[only marks, mark=*, mark options={}, mark size=1.5000pt, color=black, fill=black] table[row sep=crcr]{%
x	y\\
1.92869350327622	1.49905779305143\\
1.86426601607502	1.27202053253219\\
2.02860289157014	0.987204342571864\\
2.19931357876995	0.50281345935193\\
2.24602900099635	-0.212497810220776\\
1.96960315281106	-1.10021134928796\\
1.19541521245623	-1.9133175540128\\
-1.06078481506214e-16	-2.25605886285471\\
-1.19541521245623	-1.9133175540128\\
-1.96960315281106	-1.10021134928796\\
-2.24602900099635	-0.212497810220776\\
-2.19931357876995	0.50281345935193\\
-2.02860289157014	0.987204342571864\\
-1.86426601607502	1.27202053253219\\
-1.92869350327622	1.49905779305143\\
};

\addplot[only marks, mark=*, mark options={}, mark size=0.5000pt, draw=mycolor1] table[row sep=crcr]{%
x	y\\
2.44546216356167	1.92401390347225\\
2.18477578102527	1.71442982187651\\
2.01599165951152	1.57529644594451\\
1.91596830044358	1.48720997479326\\
1.86563971958314	1.43378193927868\\
1.84704239896224	1.39979981685965\\
1.84400373172941	1.3725374507388\\
1.84555148292548	1.34463253119805\\
1.84865972279448	1.31501588768554\\
1.85677611214838	1.28592744519487\\
1.8744960489332	1.25831703730664\\
1.90232361631342	1.22946295566852\\
1.93503647023606	1.19453850925475\\
1.96479221409711	1.15077352485146\\
1.98709370591518	1.10017168787972\\
2.00455209052859	1.04754884536817\\
2.02453826865929	0.995804314647733\\
2.0528571535252	0.943072133898453\\
2.08883444624173	0.883978159095303\\
2.12533332130364	0.814196047023674\\
2.1540577513568	0.734841567945087\\
2.17248967305058	0.651870635846663\\
2.18595730350854	0.570261411718078\\
2.20253914524377	0.489082895673643\\
2.22575889580023	0.401972435287625\\
2.25112065493867	0.302259876032293\\
2.26909784661526	0.189280350064915\\
2.27291540577883	0.0702687098943889\\
2.26442466003377	-0.0458312816235888\\
2.25173293230815	-0.156943410563192\\
2.24102045052601	-0.269740195159087\\
2.23037664624412	-0.393877599303603\\
2.21054374847159	-0.533458967035116\\
2.17251736542433	-0.681762704586014\\
2.11599599083408	-0.825509662284249\\
2.04943655938722	-0.956762399978082\\
1.98105180931508	-1.07972889837203\\
1.91039342837858	-1.20582103118716\\
1.82783858832733	-1.34300141486239\\
1.72251126963558	-1.48758591399071\\
1.59331690192751	-1.62520711774139\\
1.4516462009896	-1.74280521357187\\
1.3105411428912	-1.84013296346617\\
1.1722344608951	-1.92775702147865\\
1.0264612921456	-2.01543090099649\\
0.859457864403202	-2.10243606682012\\
0.667444483438937	-2.17695887649759\\
0.463422102035756	-2.2262798765568\\
0.266954219809759	-2.24910092967614\\
0.0864162678228271	-2.25554087997\\
-0.0864162678228274	-2.25554087997\\
-0.266954219809759	-2.24910092967614\\
-0.463422102035756	-2.2262798765568\\
-0.667444483438938	-2.17695887649759\\
-0.859457864403203	-2.10243606682012\\
-1.0264612921456	-2.01543090099649\\
-1.1722344608951	-1.92775702147865\\
-1.3105411428912	-1.84013296346617\\
-1.4516462009896	-1.74280521357187\\
-1.59331690192751	-1.62520711774139\\
-1.72251126963558	-1.48758591399071\\
-1.82783858832733	-1.34300141486239\\
-1.91039342837858	-1.20582103118716\\
-1.98105180931508	-1.07972889837203\\
-2.04943655938722	-0.956762399978082\\
-2.11599599083408	-0.825509662284249\\
-2.17251736542433	-0.681762704586013\\
-2.21054374847159	-0.533458967035115\\
-2.23037664624412	-0.393877599303603\\
-2.24102045052601	-0.269740195159086\\
-2.25173293230815	-0.156943410563192\\
-2.26442466003377	-0.0458312816235884\\
-2.27291540577883	0.0702687098943891\\
-2.26909784661526	0.189280350064915\\
-2.25112065493867	0.302259876032293\\
-2.22575889580023	0.401972435287626\\
-2.20253914524377	0.489082895673643\\
-2.18595730350854	0.570261411718078\\
-2.17248967305058	0.651870635846663\\
-2.1540577513568	0.734841567945087\\
-2.12533332130364	0.814196047023674\\
-2.08883444624173	0.883978159095303\\
-2.0528571535252	0.943072133898453\\
-2.02453826865929	0.995804314647733\\
-2.00455209052859	1.04754884536817\\
-1.98709370591518	1.10017168787972\\
-1.96479221409711	1.15077352485146\\
-1.93503647023606	1.19453850925475\\
-1.90232361631342	1.22946295566852\\
-1.8744960489332	1.25831703730664\\
-1.85677611214838	1.28592744519487\\
-1.84865972279448	1.31501588768554\\
-1.84555148292548	1.34463253119805\\
-1.84400373172941	1.3725374507388\\
-1.84704239896224	1.39979981685965\\
-1.86563971958314	1.43378193927868\\
-1.91596830044358	1.48720997479326\\
-2.01599165951152	1.57529644594451\\
-2.18477578102527	1.71442982187651\\
-2.44546216356167	1.92401390347225\\
};

\end{axis}

\end{tikzpicture}%
\end{minipage}
\begin{minipage}{0.49\textwidth}
%
%
\begin{tikzpicture}

\begin{axis}[%
width=\figurewidth,
height=\figureheight,
at={(0.772in,0.501in)},
scale only axis,
xmin=-1,
xmax=1,
xlabel style={font=\color{white!16!black}},
xlabel={$x$},
ymin=-2.5,
ymax=2.5,
ylabel style={font=\color{white!15!black}},
ylabel={$\chi_\ell(x)$},
ylabel near ticks, ylabel shift={-5pt},
xlabel near ticks, xlabel shift={-2pt},
axis background/.style={fill=white},
title style={font=\bfseries},
title={Out-of-sample extension},
axis x line*=bottom,
axis y line*=left,
tick label style={font=\tiny},
]
\addplot[only marks, mark=*, mark options={}, mark size=1.5000pt, color=black, fill=black] table[row sep=crcr]{%
x	y\\
-0.933333333333333	-1.92869350318546\\
-0.8	-1.86426601601661\\
-0.666666666666667	-2.02860289155183\\
-0.533333333333333	-2.19931357877578\\
-0.4	-2.24602900098888\\
-0.266666666666667	-1.96960315276375\\
-0.133333333333333	-1.19541521239931\\
0	-6.25528620078742e-16\\
0.133333333333333	1.19541521239931\\
0.266666666666667	1.96960315276375\\
0.4	2.24602900098888\\
0.533333333333333	2.19931357877578\\
0.666666666666667	2.02860289155184\\
0.8	1.86426601601661\\
0.933333333333333	1.92869350318546\\
};

\addplot[only marks, mark=*, mark options={}, mark size=1.5000pt, color=black, fill=black] table[row sep=crcr]{%
x	y\\
-0.933333333333333	-1.4990577931682\\
-0.8	-1.27202053261781\\
-0.666666666666667	-0.987204342609478\\
-0.533333333333333	-0.502813459326416\\
-0.4	0.212497810299772\\
-0.266666666666667	1.10021134937266\\
-0.133333333333333	1.91331755404837\\
0	2.25605886285471\\
0.133333333333333	1.91331755404837\\
0.266666666666667	1.10021134937266\\
0.4	0.212497810299772\\
0.533333333333333	-0.502813459326416\\
0.666666666666667	-0.987204342609477\\
0.8	-1.2720205326178\\
0.933333333333333	-1.4990577931682\\
};

\addplot[only marks, mark=*, mark options={}, mark size=0.5000pt, draw=blue] table[row sep=crcr]{%
x	y\\
-0.99	-2.44546216344736\\
-0.97	-2.18477578092393\\
-0.95	-2.01599165941919\\
-0.93	-1.91596830035757\\
-0.91	-1.86563971950188\\
-0.89	-1.84704239888516\\
-0.87	-1.84400373165663\\
-0.85	-1.84555148285737\\
-0.83	-1.84865972273122\\
-0.81	-1.85677611208997\\
-0.79	-1.87449604887977\\
-0.77	-1.90232361626541\\
-0.75	-1.93503647019411\\
-0.73	-1.96479221406166\\
-0.71	-1.98709370588607\\
-0.69	-2.0045520905052\\
-0.67	-2.02453826864096\\
-0.65	-2.05285715351157\\
-0.63	-2.08883444623272\\
-0.61	-2.12533332129913\\
-0.59	-2.15405775135627\\
-0.57	-2.1724896730531\\
-0.55	-2.18595730351309\\
-0.53	-2.2025391452495\\
-0.51	-2.22575889580638\\
-0.49	-2.25112065494442\\
-0.47	-2.26909784661959\\
-0.45	-2.2729154057807\\
-0.43	-2.2644246600324\\
-0.41	-2.25173293230304\\
-0.39	-2.24102045051665\\
-0.37	-2.23037664622975\\
-0.35	-2.21054374845129\\
-0.33	-2.17251736539745\\
-0.31	-2.11599599080064\\
-0.29	-2.04943655934785\\
-0.27	-1.98105180927056\\
-0.25	-1.91039342832951\\
-0.23	-1.82783858827428\\
-0.21	-1.72251126957947\\
-0.19	-1.59331690186976\\
-0.17	-1.45164620093178\\
-0.15	-1.31054114283466\\
-0.13	-1.1722344608411\\
-0.11	-1.02646129209574\\
-0.09	-0.859457864359755\\
-0.07	-0.667444483404353\\
-0.0499999999999999	-0.463422102011528\\
-0.03	-0.266954219795826\\
-0.01	-0.0864162678183345\\
0.01	0.0864162678183334\\
0.03	0.266954219795825\\
0.0499999999999999	0.463422102011527\\
0.07	0.667444483404353\\
0.09	0.859457864359755\\
0.11	1.02646129209574\\
0.13	1.1722344608411\\
0.15	1.31054114283466\\
0.17	1.45164620093178\\
0.19	1.59331690186976\\
0.21	1.72251126957947\\
0.23	1.82783858827428\\
0.25	1.91039342832951\\
0.27	1.98105180927056\\
0.29	2.04943655934785\\
0.31	2.11599599080064\\
0.33	2.17251736539745\\
0.35	2.21054374845129\\
0.37	2.23037664622975\\
0.39	2.24102045051665\\
0.41	2.25173293230304\\
0.43	2.2644246600324\\
0.45	2.2729154057807\\
0.47	2.26909784661959\\
0.49	2.25112065494442\\
0.51	2.22575889580638\\
0.53	2.2025391452495\\
0.55	2.18595730351309\\
0.57	2.1724896730531\\
0.59	2.15405775135627\\
0.61	2.12533332129913\\
0.63	2.08883444623272\\
0.65	2.05285715351157\\
0.67	2.02453826864096\\
0.69	2.0045520905052\\
0.71	1.98709370588607\\
0.73	1.96479221406166\\
0.75	1.93503647019411\\
0.77	1.90232361626541\\
0.79	1.87449604887977\\
0.81	1.85677611208998\\
0.83	1.84865972273122\\
0.85	1.84555148285737\\
0.87	1.84400373165663\\
0.89	1.84704239888516\\
0.91	1.86563971950188\\
0.93	1.91596830035757\\
0.95	2.01599165941919\\
0.97	2.18477578092393\\
0.99	2.44546216344736\\
};

\addplot[only marks, mark=*, mark options={}, mark size=0.5000pt, draw=red] table[row sep=crcr]{%
x	y\\
-0.99	-1.92401390361754\\
-0.97	-1.71442982200564\\
-0.95	-1.57529644606267\\
-0.93	-1.48720997490407\\
-0.91	-1.43378193938441\\
-0.89	-1.39979981696135\\
-0.87	-1.37253745083657\\
-0.85	-1.34463253129154\\
-0.83	-1.31501588777448\\
-0.81	-1.2859274452792\\
-0.79	-1.25831703738622\\
-0.77	-1.22946295574281\\
-0.75	-1.1945385093227\\
-0.73	-1.150773524912\\
-0.71	-1.1001716879323\\
-0.69	-1.04754884541292\\
-0.67	-0.995804314684997\\
-0.65	-0.943072133928123\\
-0.63	-0.883978159116589\\
-0.61	-0.814196047035447\\
-0.59	-0.734841567946655\\
-0.57	-0.651870635838274\\
-0.55	-0.570261411700615\\
-0.53	-0.489082895647843\\
-0.51	-0.40197243525356\\
-0.49	-0.302259875989499\\
-0.47	-0.189280350013067\\
-0.45	-0.0702687098339793\\
-0.43	0.045831281691222\\
-0.41	0.156943410636496\\
-0.39	0.269740195236849\\
-0.37	0.393877599384974\\
-0.35	0.533458967119235\\
-0.33	0.68176270467168\\
-0.31	0.825509662369975\\
-0.29	0.956762400062417\\
-0.27	1.07972889845372\\
-0.25	1.2058210312649\\
-0.23	1.34300141493459\\
-0.21	1.48758591405568\\
-0.19	1.625207117798\\
-0.17	1.74280521362003\\
-0.15	1.84013296350643\\
-0.13	1.92775702151148\\
-0.11	2.01543090102189\\
-0.09	2.10243606683788\\
-0.07	2.17695887650819\\
-0.0499999999999999	2.22627987656184\\
-0.03	2.24910092967779\\
-0.01	2.25554087997017\\
0.01	2.25554087997017\\
0.03	2.24910092967779\\
0.0499999999999999	2.22627987656184\\
0.07	2.17695887650819\\
0.09	2.10243606683788\\
0.11	2.01543090102189\\
0.13	1.92775702151148\\
0.15	1.84013296350643\\
0.17	1.74280521362003\\
0.19	1.625207117798\\
0.21	1.48758591405568\\
0.23	1.34300141493459\\
0.25	1.2058210312649\\
0.27	1.07972889845372\\
0.29	0.956762400062417\\
0.31	0.825509662369975\\
0.33	0.681762704671681\\
0.35	0.533458967119235\\
0.37	0.393877599384974\\
0.39	0.269740195236849\\
0.41	0.156943410636496\\
0.43	0.0458312816912222\\
0.45	-0.0702687098339787\\
0.47	-0.189280350013066\\
0.49	-0.302259875989498\\
0.51	-0.40197243525356\\
0.53	-0.489082895647842\\
0.55	-0.570261411700615\\
0.57	-0.651870635838273\\
0.59	-0.734841567946655\\
0.61	-0.814196047035447\\
0.63	-0.883978159116589\\
0.65	-0.943072133928123\\
0.67	-0.995804314684996\\
0.69	-1.04754884541292\\
0.71	-1.1001716879323\\
0.73	-1.150773524912\\
0.75	-1.1945385093227\\
0.77	-1.22946295574281\\
0.79	-1.25831703738622\\
0.81	-1.2859274452792\\
0.83	-1.31501588777448\\
0.85	-1.34463253129154\\
0.87	-1.37253745083657\\
0.89	-1.39979981696135\\
0.91	-1.43378193938441\\
0.93	-1.48720997490407\\
0.95	-1.57529644606267\\
0.97	-1.71442982200564\\
0.99	-1.92401390361754\\
};

\end{axis}
\end{tikzpicture}%
\end{minipage}
\caption{Illustration of the extrapolation of the SDP embedding for a sparser sampling of the interval $[-1,1]$, which accompanies Figure~\ref{Fig:Toy1}. The large black points denote training points while the smaller colored dots are out-of-sample points. This indicates the usefulness of the out-of-sample extension. On the right-hand side, red points correspond to $\ell=1$ and blue points to $\ell=2$.\label{Fig:Toy2}}
\end{figure}

\subsubsection{Kernelized variational problem with smoothness assumptions\label{sec:kernelized_intro}}
The construction described hereabove does not make assumptions on the smoothness of the integral operator in~\eqref{eq:FormalSDP}. In particular, establishing a connection between~\eqref{eq:FormalSDP} and its discretized counterpart~\eqref{eq:SDP} is non-trivial. It is therefore advantageous to make additional assumptions on the integral kernel $b(x,y)$ of the estimated nuclear operator, in order to leverage the well-known framework of estimation in a Reproducing Kernel Hilbert Space. Indeed, this setting is convenient to analyse the discretization error thanks to the representer of~\cite{MarteauFerey}. Our discussion of the kernelized variational problem relies on several key techniques of~\cite{Rudi2020global}, which addresses a different type of problem.

For defining the kernelized problem, we consider that $X$ is an open $\ell_2$-ball of radius $R>0$ in $\mathbb{R}^d$, so that the domain's boundary is `smooth'. Denote by $k(x,y)$ the kernel of a RKHS $\mathcal{H}_k$, that we assume to be strictly positive definite, and let $\phi(x) = k(x,\cdot)\in \mathcal{H}_k$ be the canonical feature map, so that $\langle \phi(x), \phi(y)\rangle = k(x,y)$. Also, we assume that this kernel is continuous and uniformly bounded: $k(x,x)\leq \kappa^2$ for all $x\in X$ and with $\kappa>0$. It is then common,
see e.g.~\cite[Section~3]{JMLR:v11:rosasco10a}, to define the restriction operator $S:\mathcal{H}_k \to L^2(X)$ as
$(S g)(x) =  g(x)/\sqrt{|X|}$ whereas its adjoint $S^* : L^2(X)\to \mathcal{H}_k$ is given by $S^*  h = \int_X h(x) \phi(x) \rmd x/\sqrt{|X|}$. Thus, the operator $SS^*$ belongs to $\mathcal{S}(L^2(X))$ and its integral kernel is $k(x,y)$. This remark motivates the following restriction: we can choose to estimate an operator $B \in \mathcal{S}(L^2(X))$ of the form
\[
  B = S \mathbb{B} S^*, \text{ where } \mathbb{B}\in \mathcal{S}(\mathcal{H}_k) \text{ is such that } \mathbb{B}\succeq 0.
\]
By using the definition of the restriction operator, we see that $B$ has the following integral kernel 
$b(x,y) = \langle \phi(x), \mathbb{B} \phi(y) \rangle,$ which is well-defined for all $x$ and $y$ in $X$. 

\paragraph{Smoothed variational problem} 
Now, we consider that  $(\mathcal{H}_k, \langle \cdot, \cdot \rangle)$ is the Sobolev space $W_2^s(X)$ with $s> d/2$, as proposed for instance in~\cite{JMLR:v11:rosasco10a}. First, for an integer $m$ such that $0<m< s-d/2$, we know that the space $W_2^s(X)$ is embedded in $\mathcal{C}^m(X)$, as shown in~\cite[Proposition~1]{Rudi2020global}. More precisely, we can associate to each element of $W_2^s(X)$ a representative in $\mathcal{C}^m(X)$. Thus, we can define a kernelized variational problem 
\begin{equation}
  \sup_{\mathbb{B}\in \mathcal{S}(\mathcal{H}_k)} f(\mathbb{B})\triangleq \Tr(S\mathbb{A}S^* S \mathbb{B} S^*) \text{ s.t. } \mathbb{B} \succeq 0 \text{ and } \langle \phi(x), \mathbb{B} \phi(x) \rangle = d(x)\ \text{ for all } x\in X,\tag{$\text{kVP}$}\label{eq:KernelizedFormalSDP}
\end{equation}
where we assume here that $d(x)$ can be written as $d(x) = \sum_{j=1}^p w_j(x)^2$ with $p$ a finite integer, for all $x\in X$ with $w_j\in W_2^s(X)$ for all $1\leq j\leq d$. The latter assumption makes sure that there exists a \emph{psd} finite rank $\mathbb{B}_\star \in \mathcal{S}(\mathcal{H}_k)$ such that $d(x) = \langle \phi(x), \mathbb{B}_\star \phi(x)\rangle$ for all $x\in X$; see the proof of Corollary~1 in~\cite{Rudi2020global}.
Next, to have an analogue of $\bar{A}$ in~\eqref{eq:FormalSDP}, we choose 
\[
\mathbb{A} = \mathbb{I}_{\mathcal{H}_k} - u\otimes\bar{u} \text{ with } \|u\|_{\mathcal{H}_k} = 1,
\] where the second term is defined as follows $(u\otimes\bar{u})g = \langle u, g\rangle u$ for all $g\in \mathcal{H}_k$. Still in analogy with~\eqref{eq:FormalSDP}, we choose $u\in \mathcal{H}_k$ such that $u(x)>0$ for all $x\in X$.
Hence, we have the integral operator
\[
  S \mathbb{A} S^* h (x)= \frac{1}{|X|}\int_X  \Big(k(x,y) - u(x) u(y)\Big) h(y) \rmd y, 
\]
By construction, $\mathbb{A}$ is the projector onto $u^\perp$ in the RKHS, and, therefore, $0\preceq\mathbb{A}\preceq \mathbb{I}_{\mathcal{H}_k}$.

\paragraph{Discretized objective}
An additional advantage of this kernelized formulation is that it allows for a natural discretization. Indeed, given a discrete set $\widehat{X} = \{x_1, \dots , x_n\}$, we can define a discrete analogue of the restriction operator $S$
as $S_n:\mathcal{H}_k \to \mathbb{R}^n$  such that
$S_n g = (1/\sqrt{n})[g(x_1) , \dots , g(x_n)]^\top.$ Its adjoint is given by $S_n^* \mathbf{v} = (1/\sqrt{n}) \sum_{i=1}^n \mathbf{v}_i \phi(x_i)$ for all $\mathbf{v}\in \mathbb{R}^n$. With the help of these definitions, the kernel matrix writes  $ S_n S_n^* = (1/n) \mathbf{K}$.
In layman's terms, if the $\{x_1, \dots , x_n\}$ are sampled independently from the uniform probability measure on $X$ and if $n$ is large enough, the event 
$
  \|S^* S - S_n^* S_n\|_{op}\lesssim 1/\sqrt{n},
$
occurs with high probability (up to $\log$ terms). This can be shown thanks to a matrix Bernstein inequality; see e.g.~\cite[Proposition~12]{rudi2015less} or the proof of Proposition~\ref{prop:TraceApproximation} in what follows. With high probability for the same event, we have the following informal discretization error bound
\begin{equation}
  \left|\Tr\left(S \mathbb{A} S^* S \mathbb{B} S^*\right)-\Tr\left(S_n \mathbb{A} S^*_n S_n \mathbb{B}S^*_n \right)\right| \lesssim  \Tr(\mathbb{B})/\sqrt{n},\label{eq:obj_rkhs_approx}
\end{equation}
if $n$ is large enough; see Proposition~\ref{prop:TraceApproximation} for a formal statement. Thus, we consider
\begin{equation}
  \sup_{\mathbb{B}\in \mathcal{S}(\mathcal{H}_k)} f_n(\mathbb{B})\triangleq \frac{1}{n}\Tr(\overline{\mathbf{K}} S_n \mathbb{B} S^*_n) \text{ s.t. } \mathbb{B} \succeq 0 \text{ and } \langle \phi(x), \mathbb{B} \phi(x) \rangle = d(x)\ \text{ for all } x\in X.\tag{$\text{kVP-}n$}\label{eq:Reg_KernelizedFormalSDP}
\end{equation}
Above, $S_n \mathbb{B} S^*_n$ is a matrix whose element $(i,j)$ is $\frac{1}{n}\langle \phi(x_i), \mathbb{B} \phi(x_j)\rangle$ and $\overline{\mathbf{K}} = \mathbf{K} - \mathbf{u}\mathbf{u}^\top$ is a matrix with $\mathbf{u} = [u(x_1) \dots, u(x_n)]^\top$. \rev{Note that thanks to the constraints, the objective of~\eqref{eq:Reg_KernelizedFormalSDP} is still upper bounded.} The equality constraints are now discretized since solving the above optimization problem with continuous equality constraints is nontrivial.
\paragraph{Scattered constraints  and regularization} To deal with the equality constraints, we use a sampling approach, inspired by~\cite[Section 5.1]{Rudi2020global}, which requires additional regularity assumptions. The idea goes as follows. If the function $x\mapsto\langle \phi(x), \mathbb{B} \phi(x) \rangle - d(x)$ is smooth enough on $X$, restricting the equality constraints on $\widehat{X}$ yields a controlled approximation provided that $\widehat{X}$ covers $X$ well enough. A key fact is that the function $x \mapsto \langle \phi(x), \mathbb{B} \phi(x) \rangle$ belongs to $W_2^s(X)$ since $\mathbb{B}$ is positive semi-definite and nuclear, as shown in~\cite[Lemma~9]{Rudi2020global}. 
After subsampling the constraints and complementing the objective with a regularization term for some $\lambda>0$, the resulting problem reads
\begin{equation}
  \sup_{\mathbb{B}\in \mathcal{S}(\mathcal{H}_k)} f_n(\mathbb{B}) - \lambda \Tr(\mathbb{B}) \text{ subject to } \mathbb{B} \succeq 0 \text{ and } \langle \phi(x_i), \mathbb{B} \phi(x_i) \rangle = d(x_i)\ \text{ for all } i\in [n].\tag{$\text{reg-kVP-}n$}\label{eq:Reg_n_KernelizedFormalSDP}
\end{equation}
Notice that we could \emph{a priori} consider a different discrete subset of $X$ for enforcing scattered constraints; by taking the same discrete set $\widehat{X}$ for discretizing the constraints as for discretizing the objective of~\eqref{eq:KernelizedFormalSDP}, we reduce the size of the final discrete optimization problem \rev{and we make sure that the constrained objective of~\eqref{eq:Reg_n_KernelizedFormalSDP} is bounded from above}.
The extra regularization term penalizes large values of $\Tr(\mathbb{B})$ which helps to improve the bound given at the RHS of~\eqref{eq:obj_rkhs_approx}. 
The constraints discretization entails an error which can be analyzed thanks to results about functions with scattered zeros~\cite{Wendland_2004}. Recall that the domain $X$ is given here by an open $\ell_2$-ball of radius $R>0$. Still following~\cite{Rudi2020global}, if we have $d(x)- \langle \phi(x), \mathbb{B} \phi(x) \rangle = 0$ for all $x\in \widehat{X}$, then, in this case, it holds that
\begin{equation}
  |d(x)- \langle \phi(x), \mathbb{B} \phi(x) \rangle|\lesssim  h_{\widehat{X},X}^m \Big(\beta_m\Tr(\mathbb{B}) + |d|_{X,m}\Big) \text{ for all } x\in X,\label{eq:diagonal_bound_fill_distance}
\end{equation}
for some positive number $\beta_m$, provided that the fill distance 
$  h_{\widehat{X},X} = \max_{x\in X}\min_{i\in [n]}\|x_i - x\|_2, 
$
is small enough; see Proposition~\ref{prop:bound_constraints} below. The RHS of~\eqref{eq:diagonal_bound_fill_distance} is bounded with high probability by a decreasing function of $n$. We now explain how.
Let $\delta\in (0,1)$. If $n$ is large enough (see Proposition~\ref{prop:fill_distance}) and if the elements of $\widehat{X}$ are sampled independently and uniformy from $X$, then with probability at least $1-\delta$, it holds, up to a factor with a logarithmic dependence on $n/\delta$,
$h_{\widehat{X},X} \lesssim  n^{-1/d},$ 
as given in~\cite[Lemma~4]{Rudi2020global}.

\paragraph{Representer theorem and finite dimensional problem} The representer theorem of~\cite{MarteauFerey} applies to the regularized problem~\eqref{eq:Reg_n_KernelizedFormalSDP} so that the optimal operator can be found in the form
\begin{equation}
    \mathbb{B} = \sum_{i,j=1}^{n} \mathbf{C}_{ij} \phi(x_i)\otimes \overline{\phi(x_j)} \text{ for some matrix } \mathbf{C}\succeq 0,\label{eq:repB}
\end{equation}
where we used the notation $\overline{\phi(x)}g = g(x)$ for all $g\in \mathcal{H}_k$ and $x\in X$.
The representer~\eqref{eq:repB} is useful to turn the variational problem into a finite dimensional optimization problem, as we explain in what follows. First, we define the Cholesky decomposition\footnote{Note that $\mathbf{K}$ is non-singular almost surely for the Sobolev kernel.} $\mathbf{K} = \mathbf{R}^\top \mathbf{R}$. Second, we define the operator $V:\mathcal{H}_k \to \mathbb{R}^n$ as
$
V = \sqrt{n}(\mathbf{R}^{-1})^\top S_{n},
$
which satisfies $VV^* = \mathbb{I}_n$, and is such that $V^* V$ is the orthogonal projector onto the span of $ \phi(x_i)$ for all $1\leq i \leq n$. Hence, following~\cite[Lemma 2]{Rudi2020global}, a finite dimensional feature map associated to $\phi(x_i)$ writes simply
$  \bm{\Phi}_i = V \phi(x_i)\in \mathbb{R}^n$,
and note that the matrix $\bm{\Phi}$, whose $i$-th column is $\bm{\Phi}_i$, is identically equal to $\mathbf{R}$.
We do the change of variable $\mathbf{B} = \mathbf{R}\mathbf{C}\mathbf{R}^\top$, so that $\mathbb{B} = V^* \mathbf{B} V$. Consequently, the problem~\eqref{eq:Reg_n_KernelizedFormalSDP} reduces to
\begin{equation}
  \max_{\mathbf{B}\succeq 0}\frac{1}{n^2}\Tr(\overline{\mathbf{K}}\bm{\Phi}^\top \mathbf{B}\bm{\Phi}) - \lambda \Tr(\mathbf{B}) \text{ subject to } \bm{\Phi}_i^\top \mathbf{B} \bm{\Phi}_i = d(x_i) \text{ for all } i\in [n], \label{eq:discrete_kernelized}
\end{equation}
where the objective is equal to $f_n(\mathbb{B}) - \lambda \Tr(\mathbb{B})$, while the constraints are $\langle\phi(x_i), \mathbb{B} \phi(x_i) \rangle = \bm{\Phi}_i^\top \mathbf{B} \bm{\Phi}_i$ for all $i\in [n]$ for $\mathbb{B} = V^* \mathbf{B} V$.\\
\noindent Further, the out-of-sample formula is completely natural in this RKHS setting, that is, 
\[
  b(x,y) = \sum_{i,j=1}^{n} k(x,x_i)[\mathbf{R}^{-1} \mathbf{B}\mathbf{R}^{-1 \top}]_{ij} k(x_j,y) \text{ with } b(x_i,x_j) =  \bm{\Phi}_i^\top \mathbf{B} \bm{\Phi}_j, \text{ for all } i,j\in [n],
\]
by construction. Algorithmically, the finite dimensional problem can be put in a form similar to~\eqref{eq:SDP} by performing the change of variables $\mathbf{B}' = \bm{\Phi}^\top \mathbf{B}\bm{\Phi}$, so that we have specifically
\[
  \max_{\mathbf{B}'\succeq 0}\Tr\left((\overline{\mathbf{K}} - \lambda n^2 \mathbf{K}^{-1}) \mathbf{B}'\right)\text{ subject to } \mathbf{B}'_{ii} = d(x_i) \text{ for all } i\in [n]. 
\]
Thus, although the definition of $\bar{\mathbf{A}}$ differs in~\eqref{eq:SDP} with respect to $\overline{\mathbf{K}} - \lambda n^2 \mathbf{K}^{-1}$, the above discrete optimization problem also involves a subtracted kernel matrix, and the amount of subtraction can be modified by varying the parameter $\lambda>0$.

Our Theorem~\ref{thm:guaranteeSmooth} below states that the operator associated with the solution of the discrete problem achieves a good approximation of the full kernelized problem with high probability; see Section~\ref{sec:Kernelized}.

\subsection{Outline} The rest of this paper is organized as follows. In Section~\ref{sec:3}, we prove the existence of a well-defined solution of~\eqref{eq:FormalSDP} given in terms of an integral kernel of a \emph{psd} symmetric nuclear operator in the absence of smoothness assumptions. Section~\ref{sec:4} discusses the reasons why the discrete version~\eqref{eq:SDP} yields often to a dimensionality reduction, whereas the out-of-sample extension formula is also derived. 
The results related to the kernelized problem~\eqref{eq:KernelizedFormalSDP} are provided in Section~\ref{sec:Kernelized}.
Finally,  empirical case studies are reported in Section~\ref{sec:6}. The numerical method is given in the appendix together with basic elements of operator theory.

\section{Discussion of the variational problem\label{sec:3}}
To start with, we explain why the diagonal of a Mercer kernel of a \emph{psd} operator in $\mathcal{S}$ cannot vanish unless it is trivially zero, so that the diagonal constraint in~\eqref{eq:FormalSDP} is not void.

In general, the kernel of a Hilbert-Schmidt operator is an element of $L^2(X\times X)$ which is not necessarily defined pointwisely everywhere.
However, from the Mercer Theorem of Scovel and Steinwart~\cite{Steinwart2012} (Theorem~\ref{Thm:NuclearDecomposition} in the appendix), we know that for each  \emph{psd} symmetric nuclear operator $K$, there exists an integral kernel  $k_M:X\times X\to \mathbb{R}$ which is defined pointwise, with eigendecomposition
\[k_M(x,y) = \sum_{\ell\geq 1} \lambda^{(\ell)} \phi^{(\ell)}(x)\phi^{(\ell)}(y),  \text{ for all } x,y\in X.
\]
Moreover, there exists an integral formula for the trace of a symmetric nuclear operator. Indeed, the trace  of a \emph{psd} symmetric nuclear  $K$ is given by the following integral formula
\begin{equation}
    \Tr(K) =\sum_{\ell\geq 1}\lambda^{(\ell)} = \sum_{\ell\geq 1}\lambda^{(\ell)}\int_X |\phi^{(\ell)}(x)|^2\rmd x= \int_X\sum_{\ell\geq 1}\lambda^{(\ell)} |\phi^{(\ell)}(x)|^2\rmd x,\label{eq:Trace}
\end{equation}
where the integral and series have been exchanged thanks to Beppo Levi's theorem. This shows additionally that $k_M(x,x)$ is an element of $L^1(X)$, which is also defined for all $x\in X$, in light of the Mercer Theorem. Moreover, the function $k_M(x,x)$ can not vanish everywhere if $K$ is not identically zero, since its trace is  $\int_X k_M(x,x) \rmd x = \Tr(K)$, as given by~\eqref{eq:Trace}.

We now state formally our main result about the existence of a \emph{psd} symmetric nuclear operator attaining the supremum in~\eqref{eq:FormalSDP}, and introduce the proof techniques.
\begin{theorem}[Existence of an optimal nuclear operator]\label{Thm:Existence}
Let $X=[-c,c]^d$ with $c>0$. Let $\bar{A}$ be a symmetric nuclear operator and  $d:X\to \mathbb{R}_{> 0}$ be a continuous positive function in $L^1(X)$. There exists a psd symmetric nuclear operator $B_\star$ and an associated Mercer kernel $b^{\star}_M(x,y)$ such that
\begin{equation*}
\Tr(\bar{A} B_\star) =\sup_{B\in \mathcal{S}} \Tr(\bar{A} B) \text{ subject to } B \succeq 0 \text{ and } b_M(x,x) \leq d(x)\quad \text{ for almost all }x\in X,
\end{equation*}
 and $b^{\star}_M(x,x) \leq d(x)$ almost everywhere.
\end{theorem}
In order to prove Theorem~\ref{Thm:Existence}, we first introduce a useful averaging technique and give some intermediate results about weak compactness.
\subsection{Averaging kernels of nuclear operators\label{sec:Averaging}}
A key point is that the diagonal of an element of $L^2(X\times X)$ is not necessarily defined. 
In order to obtain a representative with a well-defined diagonal, we use an averaging technique developed by Brislawn in~\cite{Brislawn}. 
We recall that $X= [-c,c]^d$. Define $C_r = [-r,r]^d$. Then, the average of a locally $L^{1}$ function, namely $f\in L^{1}_{\rm loc}(\mathbb{R}^d)$, is defined as follows
\[
\mathcal{A}_r f(x) = \frac{1}{|C_r|}\int_{C_r} f(x+s)\rmd s,
\]
where $|C_r| = (2r)^d$, and $r>0$ is small enough so that $x+s\in X$ for all $s\in [-r,r]^d$.
The differentiation Theorem of Lebesgue~\cite{Brislawn} states that $\lim_{r\to 0} \mathcal{A}_r f(x)  = f(x)$ almost everywhere. More generally, the average of $f\in L^{1}_{\rm loc}(\mathbb{R}^{2d})$ is
$
\mathcal{A}_r^{(2)}f(x,y) = \frac{1}{|C_r|^2}\int_{C_r \times C_r} f(x+s,y+t)\rmd s\rmd t.
$
Notice that $C_r$ is chosen to be a cube for the following reason: a cube in $\mathbb{R}^{2d}$ is the product of two cubes in $\mathbb{R}^{d}$ with the same sides.
Hence, if $b(x,y)$ is a kernel of a \emph{psd} symmetric nuclear operator, the limit
\[
\tilde{b}(x,y) = \lim_{r\to 0}\mathcal{A}^{(2)}_r b(x,y),
\]
is defined pointwise almost everywhere on $X\times X$. Again, by applying the differentiation Theorem of Lebesgue, the kernel and the averaged kernel satisfy  $\tilde{b}(x,y) =b(x,y)$ almost everywhere on $X\times X$. Therefore, the averaging procedure is a way of choosing a pointwise representative of $b(x,y)$.
Now, we consider the diagonal elements of the averaged kernel and follow an argument by Brislawn~\cite{Brislawn}. By using a Mercer representation of $b$, we find 
\begin{equation}
\tilde{b}(x,x) = \lim_{r\to 0}(\mathcal{A}^{(2)}_r b)(x,x)= \lim_{r\to 0}\sum_{\ell\geq 1}\lambda^{(\ell)} |\mathcal{A}_r\phi^{(\ell)}(x)|^2 \myeq \sum_{\ell\geq 1}\lambda^{(\ell)} |\phi^{(\ell)}(x)|^2,\label{eq:diag_Averaged}
\end{equation}
almost everywhere on $X$. Notice that we used in $\myeq$ that the series 
$
\sum_{\ell\geq 1}\lambda^{(\ell)} |\mathcal{A}_r\phi^{(\ell)}(x)|^2
$ 
converge absolutely and uniformly with respect to $r\in [0,+\infty)$ almost everywhere on $X$. This follows from the Hardy-Littlewood Maximal Theorem, as it is explained in detail in the proof of Theorem~3.1 in~\cite{Brislawn}. To summarize this subsection in other words, we found indeed that the diagonal of the averaged kernel, as given in~\eqref{eq:diag_Averaged}, is defined almost everywhere. 
\subsection{Useful compactness result}
For the proof of Theorem~\ref{Thm:Existence}, we also need the following result which relies on a classical compactness argument.
\begin{lemma}\label{Lemma:WeakCompact}
Let $s_\star$ be the supremum in Theorem~\ref{Thm:Existence}.
There is a sequence of nuclear operators $\left(B_\ell\right)_{\ell\geq 1}$  in $\mathcal{S}$ satisfying $B_\ell\succeq 0$ and $b_\ell(x,x) \leq d(x)$ such that $\Tr(\bar{A} B_\ell) \to s_\star$ when $\ell\to+\infty$. Furthermore, there exists a subsequence $\left(B_{\ell_k}\right)_{k\geq 1}$ and a nuclear operator $B_\star\in S$ such that 
$
\Tr\left( TB_{{\ell_k}}\right) \to \Tr\left( TB_\star\right),
$
when $k\to+\infty$ for all compact operator  $T$.
\end{lemma}
\begin{proof}[Proof of Lemma \ref{Lemma:WeakCompact}]
The existence of the sequence $\left(B_\ell\right)_{\ell\geq 1}$ is a consequence of the definition of the supremum. For all integers $\ell\geq 1$, since $B_\ell\succeq 0$, the trace of $B_\ell$ is its nuclear norm: $\|B_\ell\|_{\star} = \Tr(B_\ell)$. Therefore, the sequence $\left(B_\ell\right)_{\ell\geq 1}$ is within the ball $\|B\|_{\star} \leq \int_X d(x) \rmd x $. We now use the compactness for the  weak$^*$ topology.  Indeed, the space of nuclear operators $\mathcal{B}_1(L^2(X))$ is the dual of the space of compact operators $\mathcal{B}_0(L^2(X))$ and the duality pairing $\langle\cdot,\cdot\rangle$ is given by the trace, i.e., $\langle B,T\rangle = \Tr(BT)$ for all $T\in \mathcal{B}_0(L^2(X))$ and $B\in \mathcal{B}_1(L^2(X))$.  Since $\mathcal{B}_0(L^2(X))$ is a Banach space, we know that the ball of $\mathcal{B}_0(L^2(X))^*$ is compact for the weak$^*$ topology thanks to Banach-Alaoglu Theorem. Hence, there exists a subsequence $\left(B_{\ell_k}\right)_{k\geq 1}$ and a nuclear operator $B_\star$ with $\|B_\star\|_{\star} \leq \int_X d(x)\rmd x $ such that we have
$
\lim_{k\to+\infty}\Tr\left( TB_{{\ell_k}}\right) = \Tr\left( TB_\star\right),
$
for all compact operator $T$.
\end{proof}
\subsection{Main part of the proof of Theorem~\ref{Thm:Existence}}
The proof structure goes as follows. Firstly, we show that the supremum is attained by a nuclear operator which is self-adjoint and \emph{psd}. Next, we prove that this operator admits an integral kernel which is bounded almost everywhere by an envelope function on $X\times X$. Finally, the averaging technique of Section~\ref{sec:Averaging} is used to find an integral kernel satisfying the bound on its diagonal.
\begin{proof}[Proof of Theorem~\ref{Thm:Existence}]
(Existence.) By choosing different compact operators $T$ in  Lemma~\ref{Lemma:WeakCompact}, we show different properties of $B_\star$.
Firstly, let $T$ be the Hilbert-Schmidt integral operator of kernel $t(x,y) = g(x) f(y)$ with $f,g\in L^2(X)$ and $(B_{{\ell_k}})_{k\geq 1}$ such as in Lemma~\ref{Lemma:WeakCompact}. Then, we find 
\[
\Tr\left( TB_{{\ell_k}}\right)= \langle g, B_{\ell_k}f\rangle_{L^2(X)} = \langle B_{\ell_k} g,f\rangle_{L^2(X)},
\]
since $B_{{\ell_k}}$ is symmetric. By taking the limit $k\to +\infty$ in the above expression, we find that $B_\star$ is also symmetric. In particular, by taking the previous identity with $f=g$, since $0\leq \langle f, B_{\ell_k}f\rangle_{L^2(X)}$ for all $f\in L^2(X)$, we show that $B_\star$ is \emph{psd}. Hence, $B_\star$ is a \emph{psd} symmetric nuclear operator and, in view of Theorem~\ref{Thm:NuclearDecomposition}, we know that $B_\star$ admits an integral kernel given by $b_{\star,M}(x,y)$ for all $x,y\in X$, with a well-defined diagonal $b_{\star,M}(x,x)$ for all $x\in X$.\\
(Bounding the kernel.) Consequently, we have a sequence of kernels $(b_{\ell_k})_{k}$ in  $L^2(X\times X)$ weakly convergent, that is, $\langle b_{\ell_k},t\rangle_{L^2}\xrightarrow[]{k\to +\infty} \langle b_\star,t\rangle_{L^2}$ for all $t(x,y)\in  L^2(X\times X)$.  This sequence is bounded since $\|B_{\ell_k}\|_{HS}\leq \|B_{\ell_k}\|_{*}$ and therefore we can apply the Banach-Saks Theorem. 
Indeed, we can construct the Cesaro means $\hat{b}_{m} = (1/m)\sum_{k=1}^{m} b_{n_k}$ such that we have a strong convergence $\|\hat{b}_{m}-b_\star\|_{L^2}\to 0$ for $m\to \infty$, and $\hat{b}_{m}(x,x)\leq d(x)$ almost everywhere. Naturally, $\hat{b}_{m}\in \mathcal{S}$. Since this sequence converges in $L^2(X\times X)$, there exists a subsequence $(\hat{b}_{m_k})_k$ such that we have a pointwise convergence $\hat{b}_{m_k}(x,y)\to b_\star(x,y)$ for $k\to\infty$ almost everywhere on $X\times X$. By construction, since each kernel is in an envelope such that $|\hat{b}_{m_k}(x,y)|\leq \sqrt{d(x)d(y)}$, we have by taking the limit $k\to \infty$ that $\hat{b}_\star (x,y)\leq \sqrt{d(x)d(y)}$ almost everywhere on $X\times X$.\\
(Averaging the kernel.) Now, we use the averaging techniques of Section~\ref{sec:Averaging} in order to determine a representative with a well-defined diagonal. Let us define $\tilde{b}_\star(x,y) = \lim_{r\to 0} (\mathcal{A}_r^{(2)}\hat{b}_\star)(x,y)$ for almost all $(x,y)\in X\times X$. Then, in the light of~\eqref{eq:diag_Averaged}, its diagonal is upper bounded as follows
\begin{equation*}
\tilde{b}_\star(x,x) = \lim_{r\to 0} (\mathcal{A}_r^{(2)}\hat{b}_\star)(x,x) \leq \lim_{r\to 0} \frac{1}{|C_r|^2}\int_{C_r}\int_{C_r}\sqrt{d(x+s)d(y+t)}\rmd s\rmd t = d(x)  
\end{equation*}
almost everywhere on $X$, where we used the envelope obtained hereabove. This proves the desired result.
\end{proof}

\section{SDP Embedding and out-of-sample extension\label{sec:4}}

\subsection{Intuition for the dimensionality reduction\label{sec:dim_red}}
Empirically, the solution of \eqref{eq:SDP} is often a low rank matrix. In this section, we firstly provide a theoretical motivation for this observation. 
Finding the solution of \eqref{eq:SDP} is in fact equivalent to solving a convex relaxation of a rank minimization problem in terms of the nuclear norm. Indeed, since the nuclear norm is the convex envelope of the rank, Proposition~\ref{Prop:Rank} below illustrates why in practice we can expect the optimal solution of \eqref{eq:SDP} have a low rank.
\begin{proposition}[Equivalence with a nuclear norm minimization]\label{Prop:Rank}
Consider \eqref{eq:SDP} with $\bar{\mathbf{A}}\in \mathbb{R}^{n\times n}$  a psd matrix with a maximal eigenvalue strictly smaller than $1$ and let $\mathbf{\Sigma} \in \mathbb{R}^{n\times n}$ be the invertible matrix with orthogonal columns such that $\mathbb{I}-\bar{\mathbf{A}}= \mathbf{\Sigma}\mathbf{\Sigma}^\top$. Then, the optimal solution $\mathbf{X}_\star$ of
\[
\min_{\mathbf{X}\succeq 0}\|\mathbf{X}\|_{*} ,\ {\rm subject\ to}\ \diag\Big( (\mathbf{\Sigma}^{-1})^\top \mathbf{X}\mathbf{\Sigma}^{-1}\Big) = \bm{d},
\]
has the same rank as the optimal solution $\mathbf{B}_\star$ of \eqref{eq:SDP} and is given by $\mathbf{X}_\star = \mathbf{\Sigma}^\top \mathbf{B}_\star\mathbf{\Sigma}$. 
\end{proposition}
\begin{proof}[Proof of Proposition \ref{Prop:Rank}]
Notice first that maximizing $\Tr(\bar{\mathbf{A}}\mathbf{B})$ over $\mathbf{B}\succeq 0$ such that $\diag(\mathbf{B}) = \bf{d}$ is equivalent to minimizing $\Tr\big(\mathbf{B}(\mathbb{I}- \bar{\mathbf{A}})\big)$ since $\Tr(\mathbf{B})$ is fixed by the constraints. Then, we remark that $(\mathbb{I}- \bar{\mathbf{A}})\succ 0$ because the maximal eigenvalue of $\bar{\mathbf{A}}$ is strictly smaller than $1$ by assumption. Hence, a diagonalization procedure yields $\mathbb{I}- \bar{\mathbf{A}}= \mathbf{U} \mathbf{D} \mathbf{U}^\top$ where the diagonal matrix $\mathbf{D}$ is strictly positive definite. Therefore, we can also write $\mathbb{I}- \bar{\mathbf{A}} = \mathbf{\Sigma}\mathbf{\Sigma}^\top$,  where $\mathbf{\Sigma}\in \mathbb{R}^{n\times n}$ is invertible. As a consequence, the objective of the minimization problem
\[
\min_{\mathbf{B}\succeq 0} \Tr\big(\mathbf{B}(\mathbb{I}- \bar{\mathbf{A}})\big),\ {\rm subject\ to}\ \diag(\mathbf{B})= \bf{d},
\]
can be written $\Tr\big(\mathbf{B}(\mathbb{I}- \bar{\mathbf{A}})\big) = \Tr\big(\mathbf{\Sigma}^\top\mathbf{B}\mathbf{\Sigma}\big)$, where $\mathbf{\Sigma}^\top\mathbf{B}\mathbf{\Sigma}\succeq 0$.  Then, by performing the change of variables $\mathbf{X} = \mathbf{\Sigma}^\top\mathbf{B}\mathbf{\Sigma}$, we can rephrase the minimization problem as follows
\[
\min_{\mathbf{X}\succeq 0} \Tr\big(\mathbf{X}),\ {\rm subject\ to}\ \diag\Big((\mathbf{\Sigma}^{-1})^\top \mathbf{X}\mathbf{\Sigma}^{-1}\Big)= \bf{d}.
\]
Finally, we notice that $\|\mathbf{X}\|_* = \Tr(\sqrt{\mathbf{X}\mathbf{X}^\top}) = \Tr(\mathbf{X})$ since $\mathbf{X}$ is symmetric and $\mathbf{X}\succeq 0$.
\end{proof}
\begin{figure}[ht]
\centering
\begin{minipage}{0.45\textwidth}
\centering\textbf{SDP Embedding}\\
\includegraphics[width=6.5cm, height=3cm,trim={4cm 9cm 3.cm 9.cm},clip]{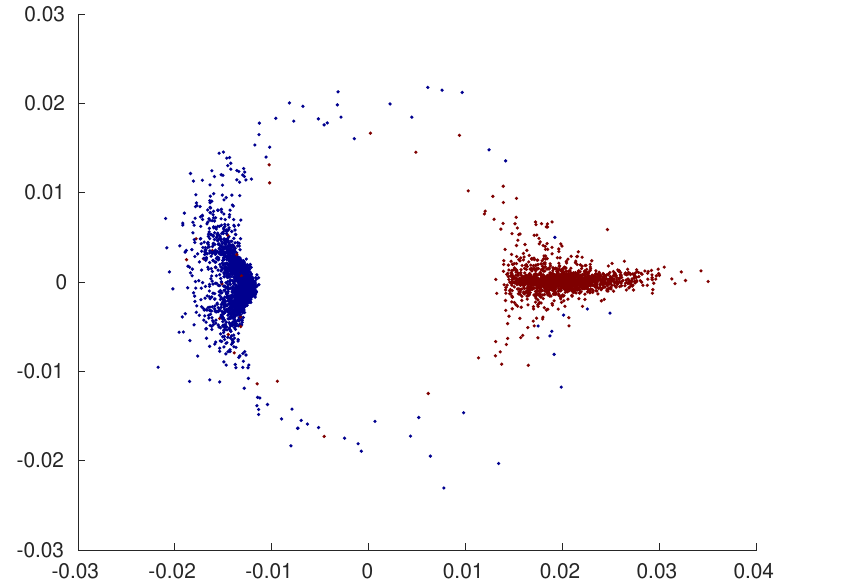}
\end{minipage}
\begin{minipage}{0.45\textwidth}
\centering\textbf{Out-of-sample extension}\\
\includegraphics[width=6.5cm, height=3cm,trim={4cm 9cm 3.cm 9.cm},clip]{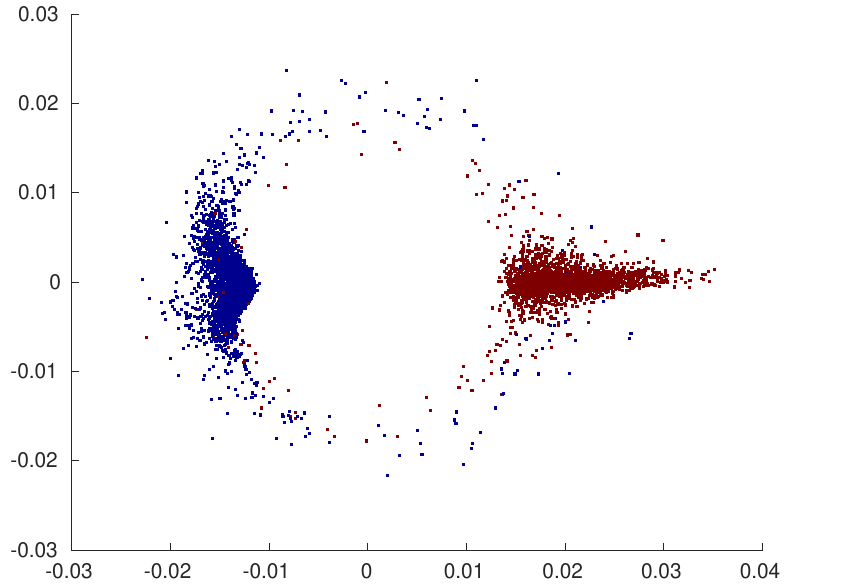}%
\end{minipage}
\caption{SDP embedding for the training set of the MNIST dataset for the digits $1$ (blue) and $4$ (red) with a bandwidth $\sigma =10$. The total number of points is here $12584$. The embedding dimension is two, i.e., $\rank(\mathbf{B}_\star) = 2$. On the left-hand side, a subsample of $3775$  points is embedded by using the extension formula~\eqref{eq:Out-of-sample2}, while the figure on the right-hand side is the embedding of a subsample of $8809$ points thanks to the out-of-sample extension. A $k$-nearest neighbours classifier was trained with $k=5$ on the first sample yielding a test error on the second sample of $1 \%$. This indicates that the out-of-sample can be useful in supervised learning.
\label{Fig:OOS_MNIST45}}
\end{figure}
\subsection{Out-of-sample extension\label{sec:oos}}
The out-of-sample extension of an eigenvector is a function given in~\eqref{eq:Out-of-sample2} which reduces to the initial vector when it is evaluated in the sample as stated in Theorem~\ref{Thm:OOS}, that we prove hereafter.
\begin{theorem}[Out-of-sample extension]\label{Thm:OOS}
Let $\bm{\chi}^{(\ell)}$ be an eigenvector of the solution of~\eqref{eq:SDP}  with non-zero eigenvalue $\lambda^{(\ell)}$ for $\ell \in [r]$, such that $\|\bm{\chi}^{(\ell)}\|_2^2 = \lambda^{(\ell)}$.
Let its extension ${\chi}_e^{(\ell)}(x)$ be given by~\eqref{eq:Out-of-sample2} where $[\bar{\bm{a}}_e(x_i)]_j = [\bar{\mathbf{A}}]_{ij}$. Then, the following properties hold
\begin{itemize}
\item[(i)] ${\chi}^{(\ell)}_e(x_i) = [\bm{\chi}^{(\ell)}]_i  \text{ for all } \ i\in [n],$
\item [(ii)] if $\bar{\bm{a}}_e(x)$ is given by~\eqref{eq:EmpiricalNormalizedKernel} and~\eqref{eq:abar}, then $\chi^{(\ell)}_e(x)$ is defined almost everywhere on $\mathbb{R}^d$.
\end{itemize}
\end{theorem}
\begin{proof}[Proof]
\noindent (i) Let $\bm{B}_\star$ be the solution of \eqref{eq:SDP}. Consider the dual certificate of \eqref{eq:SDP}, which is defined in more details in Appendix~\ref{sec:Dual},  $\mathbf{L}(\mathbf{B}) = \Diag(\bm{d})^{-1}\ddiag(\mathbf{\bar{A}}\mathbf{B})-\mathbf{\bar{A}}$ and which satisfies $\mathbf{L}(\mathbf{B}_\star)\succeq 0$ as well as $\mathbf{L}(\mathbf{B}_\star)\mathbf{B}_\star= 0$. The latter gives in components
\begin{equation}
\frac{1}{d(x_i)}[\bar{\mathbf{A}}\mathbf{B}_\star]_{ii} [\mathbf{B}_\star]_{ik}= [\bar{\mathbf{A}}\mathbf{B}_\star]_{ik} \text{ for all } i,k\in [n].\label{eq:EigenCertif2}
\end{equation}
In particular, the identity $\mathbf{L}(\mathbf{B}_\star) \bm{\chi}^{(\ell)}= 0$ holds for all eigenvector $\bm{\chi}^{(\ell)}$ of $\mathbf{B}_\star$, which gives in components:
\begin{equation}
\frac{1}{d(x_i)}[\bar{\mathbf{A}}\mathbf{B}_\star]_{ii}[\bm{\chi}^{(\ell)}]_i = [\bar{\mathbf{A}}\bm{\chi}^{(\ell)}]_i \text{ for all } i\in [n].\label{eq:EigenCertif}
\end{equation}
Let $i\in [n]$. Since $\mathbf{L}(\mathbf{B}_\star)\succeq 0$, its diagonal is non-negative, and therefore
$\frac{1}{d(x_i)}[\bar{\mathbf{A}}\mathbf{B}_\star]_{ii} \geq \bar{a}(x_i,x_i)>0.$
Then, we consider the following identity
\[
\Big([\bar{\mathbf{A}}\mathbf{B}_\star]_{ii}\Big)^2 = \sum_{j=1}^n [\bar{\mathbf{A}}\mathbf{B}_\star]_{ii}[\mathbf{B}_\star]_{ij}\bar{\mathbf{A}}_{ij}\myeq \sum_{j=1}^n d(x_i)[\bar{\mathbf{A}}\mathbf{B}_\star]_{ij}\bar{\mathbf{A}}_{ij} = d(x_i)[\bar{\mathbf{A}}\mathbf{B}_\star\bar{\mathbf{A}}]_{ii},
\]
where, in $\myeq$, we used~\eqref{eq:EigenCertif2}.
By using this identity and $[\bar{\mathbf{A}}\mathbf{B}_\star]_{ii}>0$, we find the following equality
$
[\bar{\mathbf{A}}\mathbf{B}_\star]_{ii} = \sqrt{d(x_i)}\sqrt{[\bar{\mathbf{A}}\mathbf{B}_\star\bar{\mathbf{A}}]_{ii}}$.
Then, by considering~\eqref{eq:EigenCertif}, we find
\[
[\bm{\chi}^{(\ell)}]_i = \sqrt{\frac{d(x_i)}{[\bar{\mathbf{A}}\mathbf{B}_\star\bar{\mathbf{A}}]_{ii}}}[\bar{\mathbf{A}}\bm{\chi}^{(\ell)}]_i \text{ for all } i\in[n],
\]
which finishes the first part of the proof.\\
(ii) Let $\ell\in [n]$. Consider the equation $\bar{\bm{a}}_e^\top (x) \bm{\chi}^{(\ell)} = 0$. By substituting the definition of $\bar{\bm{a}}_e(x)$, we find
\[
 \bm{\chi}^{(\ell)^\top}\bar{\bm{a}}_e(x) =   \bm{\chi}^{(\ell)^\top}(\mathbb{I} - \bm{v}^{(1)} \bm{v}^{(1)\top}) \Diag(\sqrt{\bm{m}})^{-1}\mathbf{k}(x) / \sqrt{m_e(x)} = \bm{\beta}^\top\mathbf{k}(x) / \sqrt{m_e(x)},
\]
where $\bm{\beta}$ is a suitable vector.
Hence, an equivalent condition is $f(x) =\sum_{j} \beta_j k(x_j,x) = 0$. Since $k(x,y) = e^{-\|x-y\|_2^2/\sigma^2}$, we know that $f(x)$ belongs to the Reproducing Kernel Hilbert Space (RKHS) associated to the Gaussian kernel. Hence, since the functions in this RKHS are analytic and $f$ is not identically zero, the set $\{x\in X| f(x) =0\}$ is a closed set whose interior is the empty set (cfr~\cite{RudiEtAl} and Corollary 4.44 ~\cite{Steinwart2012}).
\end{proof}
As explained in Section~\ref{sec:1}, the bound on the diagonal of the integral kernel is chosen as follows
$
d(x)= \bar{a}_e(x,x).
$
We now prove Lemma~\ref{Lemma:diagpos}, which states that this choice for $d(x)$ yields to a strictly positive function.
\begin{lemma}\label{Lemma:diagpos}
 For all $x\in X$,  we have $1/m_e(x) - m_e(x)/(\mathbf{1}^\top \mathbf{m}) >0$.
\end{lemma}
\begin{proof} Denote the Gaussian kernel by $k(x,y) = \exp(-\|x-y\|_2^2/\sigma^2)$. 
Let $\bar{\mathbf{A}} = \mathbf{A} - \bm{v}^{(1)}\bm{v}^{(1)\top}$, with $[\mathbf{A}]_{ij} = k(x_i,x_j)/\sqrt{m_e(x_i)m_e(x_j)}$ for $i,j\in [n]$ and $m(x_i) = \sum_{j=1}^{n}k(x_i,x_j)$ for all $i\in [n]$, whereas $\bm{v}^{(1)} = \sqrt{\bm{m}/(\bm{1}^\top\bm{m})}$. Since $\bar{\mathbf{A}}$ is \emph{psd}, we have $[\bar{\mathbf{A}}]_{ii}\geq 0$ for all $i\in [n]$ and as a consequence, we obtain
$
m_e(x_i)^2\leq k(x_i,x_i) \bm{1}^\top\bm{m} \text{ for all } i\in [n]
$
Now, we consider the augmented dataset $\{x_1,\dots,x_n,x\}$ with $n+1$ point and where we define $x_{n+1} = x$. Let us denote quantities relative to the augmented set by a superscript $\cdot^{(a)}$. Naturally, in that case, we also have
$m^{(a)2}_e(x_i)\leq k(x_i,x_i) \bm{1}^\top \bm{m}^{(a)}$ for all  $i\in [n+1]$
 with $m^{(a)}_e(x_i) = k(x_i,x)+m_e(x_i)$ and $\bm{1}^\top \bm{m}^{(a)}_e= \bm{1}^\top\bm{m}+2m^{(a)}_e(x)+k(x,x)$ with $x=x_{n+1}$. 
 Then, by using the above inequality with $i = n+1$, we find
 $$
 \Big(k(x,x)+m(x)\Big)^2\leq k(x ,x ) \left(\bm{1}^\top\bm{m}+2 m^{(a)}(x)+k(x,x),\right)
 $$
 with $x=x_{n+1}$. After simplification, we find $m_e(x_{n+1})^2\leq k(x ,x ) \bm{1}^\top\bm{m}$.
Since this is true for any dataset $\{x_1,\dots,x_n,x\}$, we have the desired inequality
$k(x,x)/m_e(x) - m_e(x)/(\bm{1}^\top \bm{m})>0$ for all $x\in X$.
\end{proof}

\section{Kernelized SDP Embedding\label{sec:Kernelized}}
\subsection{Objective discretization\label{sec:objectiveDiscrete}}
We begin the discussion of the kernelized problem by stating and proving the following statistical guarantees about the objective discretization.
\begin{proposition}\label{prop:TraceApproximation}
Let $\mathbb{A}$ and $\mathbb{B}$ be endomorphisms of $\mathcal{H}_k$ such that  $\mathbb{B}\in \mathcal{S}(\mathcal{H}_k)$ and $\mathbb{B}\succeq 0$, while $\mathbb{A}$ is positive semidefinite and satisfies $\|\mathbb{A}\|_{op}\leq 1$. Let $S:\mathcal{H}_k \to L^2(X)$ and $S_n:\mathcal{H}_k \to \mathbb{R}^n$ such as defined in Section~\ref{sec:kernelized_intro}, where $X$ is a bounded open set of $\mathbb{R}^d$. Let $\delta\in(0,1/2)$. Then, with probability at least $1-2\delta$, we have \begin{equation*}
    \left|\Tr\left(S\mathbb{A}S^* S \mathbb{B} S^* \right)-\Tr\left( S_n \mathbb{A} S^*_n S_n \mathbb{B}S^*_n \right)\right| \leq C_{n,\delta} \Tr(\mathbb{B}),
  \end{equation*}
  with $C_{n,\delta} = 2\kappa^2 c_{n,\delta} + c_{n,\delta}^2$ and
  $
    c_{n,\delta} = \frac{4\kappa^2 \log\left( \frac{2\kappa^2}{\lambda_{\max}(S^*S) \delta}\right)}{3n} + \sqrt{\frac{2\kappa^2 \lambda_{\max}(S^*S) \log\left( \frac{2\kappa^2}{\lambda_{\max}(S^*S) \delta}\right)}{n}}.
  $
\end{proposition}
We emphasize that if $n$ is large enough the above result yields a discretization error decaying like $1/\sqrt{n}$ if we neglect the logarithmic dependence on $\delta$.
\begin{proof}
To begin, we observe that $\mathbb{A}S^* S \mathbb{B}$ and $\mathbb{A}S^*_n S_n \mathbb{B}$ are positive semi-definite nuclear operators since $\mathbb{B}$ is nuclear while $\mathbb{A}$ and $S^* S$ are bounded and \emph{psd}. Next, we use the cyclicity property of the trace and,   by using the triangle inequality, we have 
\begin{align*}
  \left|\Tr\left(\mathbb{A}S^* S \mathbb{B} S^* S\right)-\Tr\left(\mathbb{A} S^*_n S_n \mathbb{B}S^*_n S_n \right)\right| &\leq  |\Tr\left(\mathbb{A}(S^* S - S^*_n S_n)\mathbb{B} (S^* S-S^*_n S_n)\right)|\\
  & + |\Tr\left(\mathbb{A}(S^* S - S^*_n S_n)\mathbb{B} S^*_n S_n)\right|\\
  & + |\Tr\left(\mathbb{A} S^*_n S_n\mathbb{B} (S^* S-S^*_n S_n)\right)|,
\end{align*}
where each of the terms on the RHS exists since $\mathbb{B}$ is nuclear and $\mathbb{A}$, $S^*S$ and $S^*_n S_n$ are bounded operators. First, we recall that $|\Tr(\mathbb{B}\mathbb{O})|\leq \Tr(|\mathbb{B}\mathbb{O}|)\leq \|\mathbb{O}\|_{op} \Tr(\mathbb{B})$ if $\mathbb{B}$ is nuclear and positive semidefinite, and $\mathbb{O}$ is bounded.  Now, if $\|S^*_n S_n\|_{op}\leq c_0$ and $\|S^*S-S_n^* S_n\|_{op}\leq c$, by using the submultiplicativity of the operator norm and $\|\mathbb{A}\|_{op}\leq 1$, we find
\begin{align}
  \left|\Tr\left(\mathbb{A}S^* S \mathbb{B} S^* S\right)-\Tr\left(\mathbb{A} S^*_n S_n \mathbb{B}S^*_n S_n \right)\right| \leq (c^2+2 c_0 c)\Tr(\mathbb{B}).\label{eq:structure_result}
\end{align}
Next we identify $c_0$ and $c$. First we remark that
$S_n^* S_n = \frac{1}{n}\sum_{i=1}^n \phi(x_i) \otimes \overline{\phi(x_i)}$ and $S^* S = \frac{1}{|X|}\int_{X} \phi(x) \otimes \overline{\phi(x)} \rmd x.$
Thus, $\|S_n^* S_n\|_{op}\leq \frac{1}{n}\sum_{i=1}^n k(x_i,x_i)\leq \kappa^2= c_0$ almost surely since $k(x,x)\leq \kappa^2$ for all $x\in X$ by assumption.
Next, we use a matrix Bernstein inequality for a sum of random operators to upper bound $\|S^*S-S_n^* S_n\|_{op}$.
\begin{proposition}[Proposition~12 in \cite{rudi2015less}]
  \label{p:bernstein}
Let $\calH$ be a separable Hilbert space and let $X_1, \dots , X_n$ be a sequence of independent and identically distributed self-adjoint positive random operators on $\calH$.
Assume that  $\mathbb{E} X=0$ and that there exists a real number $\ell>0$ such that
$\lambda_{\max}(X)\leq \ell$ almost surely. Let $\Sigma$ be a trace class positive operator such that $\mathbb{E}(X^2)\preceq \Sigma$. Then, for any $\delta\in (0,1)$,
\begin{equation*}
\lambda_{\max}\left(\frac{1}{n}\sum_{i=1}^n X_i\right) \leq \frac{2\ell \beta}{3n} + \sqrt{\frac{2\|\Sigma\|_{op} \beta}{n}}, \quad \text{where } \beta=\log\left(\frac{2\Tr(\Sigma)}{\|\Sigma\|_{op}\delta}\right),
\end{equation*}
with probability $1-\delta$.
If there further exists an $\ell'$ such that $\|X\|_{op}\leq \ell'$ almost surely, then, for any $\delta\in (0,1/2)$,
\begin{equation*}
\left\|\frac{1}{n}\sum_{i=1}^n X_i\right\|_{op} \leq \frac{2\ell' \beta}{3n} + \sqrt{\frac{2\|\Sigma\|_{op} \beta}{n}}, \quad \text{where } \beta=\log\left(\frac{2\Tr(\Sigma)}{\|\Sigma\|_{op}\delta}\right),
  \end{equation*}
holds with probability $1-2\delta$.
\end{proposition}
Thus, we simply define the zero-mean random variable  $X_i = \phi(x_i) \otimes \overline{\phi(x_i)} - \frac{1}{|X|}\int_X \phi(x) \otimes \overline{\phi(x)} \rmd x$ and find that $\| X \|_{op} \leq 2 \kappa^2$ holds almost surely by using a triangle inequality. It is easy to verify that 
\[
  \mathbb{E}[X_i^2]\leq \kappa^2 \mathbb{E}[\phi(x_i) \otimes \overline{\phi(x_i)}] = \kappa^2 \frac{1}{|X|}\int_X \phi(x) \otimes \overline{\phi(x)} \rmd x = \kappa^2 S^*S = \Sigma.
\]
Thus we find $\Tr(\Sigma) = \kappa^2 \int_X k(x,x) \rmd x /|X|  \leq \kappa^4$ and $\|\Sigma\|_{op} \leq \kappa^2 \lambda_{\max}(S^*S)$. By using these identifications, we have $
\beta \leq \log\left( \frac{2\kappa^2}{\lambda_{\max}(S^*S) \delta}\right)$. Then, we apply Proposition~\ref{p:bernstein} and conclude that, with probability at least $1-2\delta$,
\[
\|S^*S - S^*_n S_n\|_{op}\leq \frac{4\kappa^2 \log\left( \frac{2\kappa^2}{\lambda_{\max}(S^*S) \delta}\right)}{3n} + \sqrt{\frac{2\kappa^2 \lambda_{\max}(S^*S) \log\left( \frac{2\kappa^2}{\lambda_{\max}(S^*S) \delta}\right)}{n}} = c,
\]
where $c = c_{n,\delta}$.
We now  use the structural inequality~\eqref{eq:structure_result} and the result follows.
\end{proof}

\subsection{Constraints discretization\label{sec:constraintsDiscrete}} 
The following result is directly adapted from Theorem~4 in~\cite{Rudi2020global} (and its proof).
\begin{proposition}[see~\cite{Rudi2020global}]\label{prop:bound_constraints}
  Let $X$ be an open $\ell_2$-ball of diameter $2R$ in $\mathbb{R}^d$. Let $k$ be the kernel of the RKHS $\mathcal{H}_k = W_2^s(X)$ with $s>d/2$ and $\phi(x)$ be its canonical feature map as defined in Section~\ref{sec:kernelized_intro}. Let $0<m< s-d/2$ be an integer. Let $\widehat{X}=\{x_1,\dots, x_n\} \subset X$ such that $h_{\widehat{X},X} \leq R \min(1,\frac{1}{18(m-1)^2})$. Let $d\in \mathcal{C}^m(X)$ and assume that 
  $
    d(x_i) = \langle\phi(x_i), \mathbb{B} \phi(x_i)\rangle \text{ for all } i\in [n],
  $
  where $\mathbb{B}\succeq 0$ is a nuclear endomorphism of $\mathcal{H}_k$. Then, it holds that
  \begin{equation*}
    |d(x)- \langle \phi(x), \mathbb{B} \phi(x) \rangle|\leq \alpha_{m,d}   \left(\beta_{m}\Tr(\mathbb{B}) + |d|_{X,m}\right) h_{\widehat{X},X}^m \text{ for all } x\in X,
  \end{equation*}
  where $\alpha_{m,d}$ and $\beta_{m}$ are constants.
\end{proposition}
\begin{proposition}[Lemma 4 in~\cite{Rudi2020global}]\label{prop:fill_distance}
  Let $X\subset \mathbb{R}^d$ be a bounded set with diameter $2R$, for some $R >0$, and such that $X = \cup_{x\in S} B_r(x),$ where $S$ is a bounded subset of $\mathbb{R}^d$ for a given $r >0$. Let $\widehat{X}=\{x_1, \dots , x_n\}$ be a set of independent points sampled from the uniform distribution on $X$. When $n\geq 2(\frac{6R}{r})^d(\log\left(\frac{2}{\delta}\right)+ 2d\log\left(\frac{4R}{r}\right))$, then the following holds with probability at least $1-\delta$ : 
  \[
    h_{\widehat{X},X}\leq 11 R n^{-1/d}\left(\log\left(\frac{n}{\delta}\right)+d\log\left(\frac{2R}{r}\right)\right)^{1/d}.
  \]
\end{proposition}

\subsection{Putting all together}

 Sections~\ref{sec:objectiveDiscrete} and~\ref{sec:constraintsDiscrete} are now used to state and prove our main result about the approximation of the full problem with strong smoothness assumptions.
\begin{theorem}[Approximation of the optimal objective]\label{thm:guaranteeSmooth}
Let $\mathcal{H}_k = W_2^s(X)$ where $X$ is an open $\ell_2$-ball of radius $R$ in $\mathbb{R}^d$.  Let an integer $m$ such that $0<m<s-d/2$. Let $\mathbb{B}_{\star\star}\in \mathcal{S}(\mathcal{H}_k)$ be an optimizer of the full problem~\eqref{eq:KernelizedFormalSDP}.  Denote by $\mathbf{B}_\star$ the $n\times n$ matrix obtained by solving~\eqref{eq:discrete_kernelized} and consider $V:\mathcal{H}_k \to \mathbb{R}^n$ defined as in Section~\ref{sec:kernelized_intro}. Let $\delta\in (0,1/3)$.  Consider $C_{n,\delta}$ as given in Proposition~\ref{prop:TraceApproximation}. Then, we have:\\
(i) if $\lambda\geq 2 C_{n,\delta}$, with probability at least $1-2\delta$,
$
  |f(V^*\mathbf{B}_\star V) - f(\mathbb{B}_{\star\star})| \leq 3\lambda\Tr(\mathbb{B}_{\star\star}).
$\\
(ii) if $n$ large enough such that
 $ \frac{11}{n^{1/d}} \left(\log\left(\frac{n}{\delta}\right) +d \log(2)\right)^{\frac{1}{d}}\leq \min(1,\frac{1}{18(m-1)^2}),$ and if  $\lambda\geq 2 C_{n,\delta}$, then, with probability at least $1-\delta$, it holds
\[|d(x)- \langle \phi(x), V^*\mathbf{B}_\star V \phi(x) \rangle |\leq n^{-\frac{m}{d}} \gamma_{m,d,R} \left(\frac{3 \lambda}{2C_{n,\delta}}\beta_m \Tr(\mathbb{B}_{\star\star}) + |d|_{X,m} \right) \left(\log\left(\frac{n}{\delta}\right) +  d \log(2)\right)^{\frac{m}{d}},\]
for all $x\in X$ and for some constants $\gamma_{m,d,R}$ and $\beta_m>0$.\\
\noindent Further, with the same assumptions on $n$ and $\lambda$ in (i) and (ii),  the two bounds in (i) and (ii) hold together with probability at least $1-3\delta$.
\end{theorem}
Thus, in a nutshell, the operator $V^*\mathbf{B}_\star V$ has an objective value which is close to the optimal objective and approximately satisfies the constraints, with high probability, provided that $\lambda>0$ decays as $1/\sqrt{n}$ up to logarithmic terms.
\begin{proof}
The proof technique mainly relies on~\cite[Proof of Thm 5]{Rudi2020global}, and is also used in~\cite{Fanuel2021NonParametricDPPs}.\\
\noindent{\bf Proof of (i).} First, the full problem~\eqref{eq:KernelizedFormalSDP} has at least one feasible point since $d(x) = \sum_{j=1}^p w_j(x)^2$ with $p$ a finite integer, for all $x\in X$ with $w_j\in W_2^s(X)$ for all $1\leq j\leq d$. Indeed, $\sum_{j=1}^p w_j\otimes \overline{w_j}$ is feasible.
The objective value in~\eqref{eq:discrete_kernelized} for $\mathbf{B}_\star$ is the objective value of $V^*\mathbf{B}_\star V $ for~\eqref{eq:Reg_n_KernelizedFormalSDP}. Second, we know that there exists $\bar{\mathbf{B}}= V\mathbb{B}_{\star\star}V^*$ such that $f_n(V^*\bar{\mathbf{B}}V) = f_n(\mathbb{B}_{\star\star})$ and $\Tr(\bar{\mathbf{B}})\leq \Tr(\mathbb{B}_{\star\star})$ as a consequence of Lemma~3 in~\cite{Rudi2020global}. Hence, by the optimality of $V^*\mathbf{B}_\star V $, we find
 \[
   f_n(V^*\mathbf{B}_\star V ) - \lambda \Tr(V^*\mathbf{B}_\star V) \geq f_n(V^*\bar{\mathbf{B}} V ) - \lambda \Tr(V^*\bar{\mathbf{B}} V) \geq f_n( \mathbb{B}_{\star\star}) - \lambda \Tr(\mathbb{B}_{\star\star}).
 \]
 Hence, since $VV^* = \mathbb{I}_n$, this gives
 $
   f_n(V^*\mathbf{B}_\star V) - f_n( \mathbb{B}_{\star\star})\geq \lambda \Tr(\mathbf{B}_{\star}) - \lambda \Tr(\mathbb{B}_{\star\star}).
 $
 Now, we use Proposition~\ref{prop:TraceApproximation}, and find that the event
$  \left|f( \mathbb{B}_{\star\star}) - f_n( \mathbb{B}_{\star\star})\right| \leq C_{n,\delta} \Tr(\mathbb{B}_{\star\star}),$ with $C_{n,\delta} = (2\kappa^2 c_{n,\delta} + c_{n,\delta}^2)$,
occurs with probability $1-2\delta$. By combining this with the previous result, we find
\begin{align}
  f(\mathbb{B}_{\star\star}) - f_{n}(V^*\mathbf{B}_\star V) &= f(\mathbb{B}_{\star\star})- f_n(\mathbb{B}_{\star\star})  + f_n(\mathbb{B}_{\star\star}) - f_{n}(V^*\mathbf{B}_\star V)\nonumber\\
  & \leq C_{n,\delta} \Tr(\mathbb{B}_{\star\star}) + \lambda\left(\Tr(\mathbb{B}_{\star\star}) - \Tr(\mathbf{B}_{\star}) \right)\label{eq:key}\\
  &\leq \left( C_{n,\delta} + \lambda \right) \Tr(\mathbb{B}_{\star\star}),\label{eq:firstbound}
\end{align}
where we used $\mathbf{B}_{\star}\succeq 0$. Similarly, we have
\begin{align}
   f_{n}(V^*\mathbf{B}_\star V) - f(\mathbb{B}_{\star\star}) &= f_{n}(V^*\mathbf{B}_\star V) - f(V^*\mathbf{B}_\star V) + \underbrace{f(V^*\mathbf{B}_\star V) - f(\mathbb{B}_{\star\star})}_{\leq 0}\nonumber\\ &\leq C_{n,\delta} \Tr(V^*\mathbf{B}_\star V)= C_{n,\delta} \Tr(\mathbf{B}_\star).\label{eq:key2}
\end{align}
Using~\eqref{eq:key} with the latter inequality, we find
$
  -C_{n,\delta} \Tr(\mathbf{B}_\star) \leq C_{n,\delta} \Tr(\mathbb{B}_{\star\star}) + \lambda\left(\Tr(\mathbb{B}_{\star\star}) - \Tr(\mathbf{B}_{\star}) \right).
$
If $\lambda\geq 2 C_{n,\delta}$, this becomes $C_{n,\delta} \Tr(\mathbf{B}_\star) \leq (C_{n,\delta} + \lambda)\Tr(\mathbb{B}_{\star\star})$. Hence, by combining this with~\eqref{eq:key2} and recalling~\eqref{eq:firstbound}, we obtain, with probability at least $1-2\delta$,
\begin{equation}
  |f_{n}(V^*\mathbf{B}_\star V) - f(\mathbb{B}_{\star\star})|\leq (C_{n,\delta} + \lambda)\Tr(\mathbb{B}_{\star\star})\leq \frac{3}{2}\lambda\Tr(\mathbb{B}_{\star\star}).\label{eq:firstBound}
\end{equation}
In the light of the proof of Proposition~\ref{prop:TraceApproximation}, $|f_{n}(V^*\mathbf{B}_\star V) - f(V^*\mathbf{B}_\star V)|\leq C_{n,\delta} \Tr(V^*\mathbf{B}_\star V)$ occurs with probability at least $1-2\delta$ due to the same event as in~\eqref{eq:firstBound}.
Thus, by using a triangle inequality and the same bound on $\Tr(\mathbf{B}_\star)$  as above, we obtain, with probability at least $1-2\delta$,
\[
  |f(V^*\mathbf{B}_\star V) - f(\mathbb{B}_{\star\star})| \leq |f(V^*\mathbf{B}_\star V) - f_{n}(V^*\mathbf{B}_\star V)| +| f_{n}(V^*\mathbf{B}_\star V) - f(\mathbb{B}_{\star\star})| \leq 3 \lambda \Tr(\mathbb{B}_{\star\star}).
\]
\noindent{\bf Proof of (ii).} We simply apply Proposition~\ref{prop:bound_constraints} to  bound $|d(x)- \langle \phi(x), V^*\mathbf{B}_\star V  \phi(x) \rangle|$ in terms of the fill distance $h_{\widehat{X},X}$, and $\Tr(V^*\mathbf{B}_\star V ) = \Tr(\mathbf{B}_\star)$, if $h_{\widehat{X},X} \leq R \min(1,\frac{1}{18(m-1)^2})$.  The trace $\Tr(\mathbf{B}_\star)$ is upper bounded as above, i.e., if $\lambda\geq 2 C_{n,\delta}$, we have
\[
  C_{n,\delta} \Tr(\mathbf{B}_\star) \leq \frac{C_{n,\delta} + \lambda}{C_{n,\delta}}\Tr(\mathbb{B}_{\star\star})\leq \frac{3\lambda}{2C_{n,\delta}}\Tr(\mathbb{B}_{\star\star}).
\]
As stated in Proposition~\ref{prop:fill_distance}, the fill distance is bounded with probability at least $1-\delta$ by
\[
  h_{\widehat{X},X}\leq 11 R n^{-1/d}\left(\log\left(\frac{n}{\delta}\right)+d\log(2)\right)^{1/d}.
\]
Hence we require $ \frac{11R}{n^{1/d}} \left(\log\left(\frac{n}{\delta}\right) +d \log(2)\right)^{\frac{1}{d}}\leq R\min(1,\frac{1}{18(m-1)^2})$. By combining these results, we obtain (ii).
\end{proof}

\section{Illustrative examples of dimensionality reduction with weak smoothness assumptions\label{sec:6}}
\rev{The datasets used in the simulations are: Wine\footnote{\url{https://archive.ics.uci.edu/ml/datasets/wine}} ($n=178$, $d=13$), the digits $1$ and $4$ of the training data of
MNIST\footnote{\url{http://yann.lecun.com/exdb/mnist/}} ($n=12584$, $d =784$), the HTRU2 dataset\footnote{\url{https://archive.ics.uci.edu/ml/datasets/HTRU2}} $(n=17898, d=9)$, which is a classification benchmark ($1639$ pulsars vs $16259$ non-pulsars), and a commonly used artificial example with two half-moons\footnote{See  \texttt{twomoons} \url{https://nl.mathworks.com/help/stats/label-data-using-semi-supervised-learning-techniques.html}}($n=400$, $d=2$). Our code is available at \url{https://github.com/mrfanuel/sdp-embedding}.}

As an illustration of the out-of-sample formula, the digits $1$ and $4$ of the training set of the MNIST dataset are visualized in Figure~\ref{Fig:OOS_MNIST45}. On the left, $30$ percent of the data is embedded thanks to the SDP embedding while the out-of-sample formula is used in order to embed the remaining $70$ percent of the dataset as displayed on the right of Figure~\ref{Fig:OOS_MNIST45}. 
 In the case displayed in Figure~\ref{Fig:OutliersBandwidth}, the embedding dimension is two, since the rank of the optimal solution $\mathbf{B}_\star$ is equal to $2$. In the case of the Wine dataset given in Figure~\ref{Fig:Wine}, we observe again that the result of the SDP embedding gives still an interesting information (bottom left) when the bandwidth is chosen so small that Diffusion Maps only emphasizes the outliers (bottom right).
\begin{figure}[h]
  \centering
  \setlength\figureheight{0.15\textwidth} 
  \setlength\figurewidth{0.35\textwidth}
  \begin{subfigure}{0.45\textwidth}
  \input{Figures/SDPwine10.tikz}
  \end{subfigure}
  \hfill
  \begin{subfigure}{0.45\textwidth}
  \input{Figures/DMwine10.tikz}
  \end{subfigure}
  \begin{subfigure}{0.45\textwidth}
  \input{Figures/SDPwine1c5.tikz}
  \end{subfigure}
  \hfill
  \begin{subfigure}{0.45\textwidth}
  \input{Figures/DMwine1c5.tikz}
  \end{subfigure}
  \begin{subfigure}{0.45\textwidth}
%
%
\definecolor{mycolor1}{rgb}{0.00000,0.44700,0.74100}%
\definecolor{mycolor2}{rgb}{0.85000,0.32500,0.09800}%
\definecolor{mycolor3}{rgb}{0.92900,0.69400,0.12500}%
\definecolor{mycolor4}{rgb}{0.49400,0.18400,0.55600}%
\definecolor{mycolor5}{rgb}{0.46600,0.67400,0.18800}%
\begin{tikzpicture}

\begin{axis}[%
width=\figurewidth,
height=\figureheight,
at={(0.758in,0.481in)},
scale only axis,
xmin=0.4,
xmax=6,
ymin=0,
ymax=0.9,
axis background/.style={fill=white},
title style={font=\bfseries},
tick label style={font=\tiny},
title={Spectrum of $\mathbf{B}_\star/\Tr(\mathbf{B}_\star)$},
xlabel={$\sigma$},
]
\addplot [color=mycolor1, mark=*, mark size=1pt, mark options={solid, mycolor1}, forget plot]
  table[row sep=crcr]{%
0.5	0.11171353779996\\
0.6	0.424043614805521\\
0.7	0.765641366584053\\
0.8	0.721297801186335\\
0.9	0.690321097291475\\
1	0.658052885712732\\
1.1	0.607459978204613\\
1.2	0.576092213525147\\
1.3	0.551177270631731\\
1.4	0.550873208762133\\
1.5	0.550761978075096\\
1.6	0.559004355284948\\
1.7	0.546459544778256\\
1.8	0.537430194532044\\
1.9	0.548976444928223\\
2	0.575283902741641\\
2.1	0.606382811068074\\
2.2	0.641946589763198\\
2.3	0.670778163605604\\
2.4	0.688277640233718\\
2.5	0.701422903511918\\
2.6	0.711784858043353\\
2.7	0.719798953278321\\
2.8	0.729762741017056\\
2.9	0.737509149368668\\
3	0.743947111006156\\
3.1	0.751114748444715\\
3.2	0.758434047017136\\
3.3	0.765152170845295\\
3.4	0.771408165062316\\
3.5	0.777357171073097\\
3.6	0.78294772492327\\
3.7	0.788283684138371\\
3.8	0.793331693498962\\
3.9	0.797946658168707\\
4	0.802424224480193\\
4.1	0.806411743709882\\
4.2	0.810289851038027\\
4.3	0.813774286630242\\
4.4	0.817239609130623\\
4.5	0.820418002235923\\
4.6	0.823402375362292\\
4.7	0.826280135465155\\
4.8	0.834276663497877\\
4.9	0.831167637543227\\
5	0.838600370138656\\
5.1	0.839496159078863\\
5.2	0.843412652334032\\
5.3	0.844516553062526\\
5.4	0.846267161167363\\
5.5	0.848487957325699\\
5.6	0.850202721040027\\
5.7	0.851134263893309\\
5.8	0.852288341846431\\
5.9	0.85547530345844\\
6	0.85707924033178\\
};
\addplot [color=mycolor2, mark=*, mark size=1pt, mark options={solid, mycolor2}, forget plot]
  table[row sep=crcr]{%
0.5	0.0956747268787906\\
0.6	0.168329038178698\\
0.7	0.226700640676411\\
0.8	0.274920524908394\\
0.9	0.306041943806385\\
1	0.337109460432003\\
1.1	0.389125750897437\\
1.2	0.423289979075898\\
1.3	0.433861080949295\\
1.4	0.445550129559504\\
1.5	0.448238212659434\\
1.6	0.440576807537454\\
1.7	0.45020335221781\\
1.8	0.460217292403664\\
1.9	0.450975092744608\\
2	0.423290647009449\\
2.1	0.391166168041612\\
2.2	0.35645758522095\\
2.3	0.328036873582827\\
2.4	0.310510053832989\\
2.5	0.297312137798589\\
2.6	0.286665616694646\\
2.7	0.278177398908608\\
2.8	0.270176732686606\\
2.9	0.262450006723874\\
3	0.255186493461909\\
3.1	0.248217024421518\\
3.2	0.241527134545799\\
3.3	0.234812986416962\\
3.4	0.228572086840188\\
3.5	0.222608712526181\\
3.6	0.21702257688657\\
3.7	0.211689078222139\\
3.8	0.206644393861184\\
3.9	0.202032019919264\\
4	0.19755515756183\\
4.1	0.193570607699535\\
4.2	0.189696932561045\\
4.3	0.186209308771786\\
4.4	0.182741736156075\\
4.5	0.179565591115941\\
4.6	0.176584821246654\\
4.7	0.173707613121239\\
4.8	0.165232202977079\\
4.9	0.168820629334289\\
5	0.16094962341987\\
5.1	0.159922768213462\\
5.2	0.15621030479591\\
5.3	0.154997482114787\\
5.4	0.15320322003108\\
5.5	0.151033202681333\\
5.6	0.149385356874967\\
5.7	0.148380933672725\\
5.8	0.147295357293421\\
5.9	0.143995060211751\\
6	0.142575556076531\\
};
\addplot [color=mycolor3, mark=*, mark size=1pt, mark options={solid, mycolor3}, forget plot]
  table[row sep=crcr]{%
0.5	0.0553158576285013\\
0.6	0.0569347540953612\\
0.7	0.0017883528030328\\
0.8	0.000877178868209799\\
0.9	0.000537720805398824\\
1	0.0024246922525829\\
1.1	0.000903736218337667\\
1.2	0.000201813850398902\\
1.3	0.0131760817557373\\
1.4	0.00110655381119589\\
1.5	0.000771222056245809\\
1.6	0.000270742556067016\\
1.7	0.00144880897568101\\
1.8	0.000920716182858566\\
1.9	3.46949258660424e-05\\
2	0.000745934894018999\\
2.1	0.00199628350018917\\
2.2	0.0011947122022628\\
2.3	0.00103257443179795\\
2.4	0.00107316881759195\\
2.5	0.00122282224855082\\
2.6	0.00153471691652222\\
2.7	0.00195734037942105\\
2.8	6.01147214134374e-05\\
2.9	4.02167291941598e-05\\
3	0.000851095733213014\\
3.1	0.000654510025821927\\
3.2	3.60920136515062e-05\\
3.3	3.24556263108703e-05\\
3.4	1.85662548350719e-05\\
3.5	3.27697465546823e-05\\
3.6	2.86376326937904e-05\\
3.7	2.65289500228337e-05\\
3.8	2.32164423643849e-05\\
3.9	2.08475194668119e-05\\
4	2.01669193033375e-05\\
4.1	1.73940910312104e-05\\
4.2	1.28956861171079e-05\\
4.3	1.62702444262211e-05\\
4.4	1.85079724666273e-05\\
4.5	1.61993053540262e-05\\
4.6	1.26357882784766e-05\\
4.7	1.20328205228641e-05\\
4.8	0.000418351839384629\\
4.9	1.16707799693939e-05\\
5	0.000399789484011882\\
5.1	0.000544347005554416\\
5.2	0.000336294008943835\\
5.3	0.000450367500892611\\
5.4	0.000499829052606973\\
5.5	0.000445880425616845\\
5.6	0.000380613318946635\\
5.7	0.00046508607162062\\
5.8	0.000389688773997182\\
5.9	0.000516985014924258\\
6	0.00032992685050568\\
};
\addplot [color=mycolor4, mark=*, mark size=1pt, mark options={solid, mycolor4}, forget plot]
  table[row sep=crcr]{%
0.5	0.0514778312041747\\
0.6	0.0338139945588336\\
0.7	0.00116099477410988\\
0.8	0.000528670872501245\\
0.9	0.000501539069381118\\
1	0.000438937045028576\\
1.1	0.000489299933972818\\
1.2	0.000115073318398308\\
1.3	0.00045588405858225\\
1.4	0.000924727368081234\\
1.5	0.000119185477038905\\
1.6	8.51427128255952e-05\\
1.7	0.000985322615857289\\
1.8	0.000625680786911475\\
1.9	1.1619567016811e-05\\
2	0.00033091811132768\\
2.1	0.000371489968671274\\
2.2	0.00034486615333271\\
2.3	9.90055570808545e-05\\
2.4	0.000114725744458906\\
2.5	2.19732760939454e-05\\
2.6	1.15380401978786e-05\\
2.7	4.94895039763522e-05\\
2.8	2.89134302977892e-07\\
2.9	5.5169783557439e-07\\
3	1.02683635662292e-05\\
3.1	7.98376071881666e-06\\
3.2	2.65279296308061e-06\\
3.3	2.17840090273525e-06\\
3.4	8.69693911929525e-07\\
3.5	1.02018214793477e-06\\
3.6	8.57835980420398e-07\\
3.7	4.82496571489882e-07\\
3.8	4.7608210855164e-07\\
3.9	2.73599792641459e-07\\
4	2.6315783470096e-07\\
4.1	1.43572816860719e-07\\
4.2	2.13595281795011e-07\\
4.3	1.00609972985942e-07\\
4.4	1.02772938317464e-07\\
4.5	1.21646888769016e-07\\
4.6	9.78169728206031e-08\\
4.7	1.7321629321248e-07\\
4.8	6.0696023390436e-05\\
4.9	5.45342699798032e-08\\
5	4.41566566315047e-05\\
5.1	3.2174796632829e-05\\
5.2	3.71201716768105e-05\\
5.3	3.30703057197528e-05\\
5.4	2.77129773912454e-05\\
5.5	3.13523560889102e-05\\
5.6	3.03713348345783e-05\\
5.7	1.88811120499988e-05\\
5.8	2.59671627414302e-05\\
5.9	1.21269030498555e-05\\
6	1.47270177404768e-05\\
};
\addplot [color=mycolor5, mark=*, mark size=1pt, mark options={solid, mycolor5}, forget plot]
  table[row sep=crcr]{%
0.5	0.0469055196608219\\
0.6	0.0268462251109569\\
0.7	0.000614158983142027\\
0.8	0.000387231562170721\\
0.9	0.000385622209160398\\
1	0.00029432476652299\\
1.1	0.000365216958008282\\
1.2	7.7724665408831e-05\\
1.3	0.000349887633753862\\
1.4	0.00045407345044558\\
1.5	4.92305555157758e-05\\
1.6	3.50389710093468e-05\\
1.7	0.000510132030882784\\
1.8	0.000452196943478792\\
1.9	1.43254548832018e-06\\
2	0.000243235560719556\\
2.1	8.21297948422101e-05\\
2.2	3.51321204039499e-05\\
2.3	2.64907898779137e-05\\
2.4	1.30345530052155e-05\\
2.5	1.51926597061449e-05\\
2.6	1.91899092846043e-06\\
2.7	1.24884725530336e-05\\
2.8	1.08201531906853e-07\\
2.9	6.9025641397645e-08\\
3	4.6439062536794e-06\\
3.1	5.34162977552166e-06\\
3.2	6.76283002059359e-08\\
3.3	2.00441402430118e-07\\
3.4	3.06248211565652e-07\\
3.5	3.21058653034099e-07\\
3.6	1.99986970201674e-07\\
3.7	2.25488981935705e-07\\
3.8	2.19773141355978e-07\\
3.9	1.95882041587159e-07\\
4	1.86963526823943e-07\\
4.1	1.08759893427713e-07\\
4.2	1.06439678128773e-07\\
4.3	2.86761669433594e-08\\
4.4	4.33363344797468e-08\\
4.5	8.54081632178471e-08\\
4.6	6.96655453598255e-08\\
4.7	4.52989049932748e-08\\
4.8	1.16270158201934e-05\\
4.9	7.68587711220156e-09\\
5	5.45600106519416e-06\\
5.1	3.78813236027311e-06\\
5.2	3.069773957144e-06\\
5.3	1.95701219982693e-06\\
5.4	1.57433481997275e-06\\
5.5	1.23505582037242e-06\\
5.6	5.96270268761356e-07\\
5.7	4.56628638900546e-07\\
5.8	3.28047667715915e-07\\
5.9	2.95531496228487e-07\\
6	3.67774204143007e-07\\
};
\end{axis}
\end{tikzpicture}%
  \end{subfigure}
  \hfill
  \begin{subfigure}{0.45\textwidth}
%
%
\definecolor{mycolor1}{rgb}{0.00000,0.44700,0.74100}%
\definecolor{mycolor2}{rgb}{0.85000,0.32500,0.09800}%
\definecolor{mycolor3}{rgb}{0.92900,0.69400,0.12500}%
\begin{tikzpicture}

\begin{axis}[%
width=\figurewidth,
height=\figureheight,
at={(0.772in,0.481in)},
scale only axis,
xmin=1.5,
xmax=12,
ymin=0,
ymax=1,
tick label style={font=\tiny},
axis background/.style={fill=white},
title style={font=\bfseries},
title={Spectrum of $\bar{\mathbf{A}}$},
xlabel={$\sigma$},
]
\addplot [color=mycolor1, mark=*, mark size=1pt, mark options={solid, mycolor1}, forget plot]
  table[row sep=crcr]{%
1.5	0.998426114836858\\
1.6	0.996271271998048\\
1.7	0.992157343192957\\
1.8	0.984930277072668\\
1.9	0.973043577939594\\
2	0.954599780108485\\
2.1	0.92887689326661\\
2.2	0.911993133471217\\
2.3	0.893898857467954\\
2.4	0.874208124414888\\
2.5	0.85309734914895\\
2.6	0.830786964190463\\
2.7	0.807519065997141\\
2.8	0.783542635853523\\
2.9	0.759101568990437\\
3	0.734425470072221\\
3.1	0.709723239457899\\
3.2	0.685179224264412\\
3.3	0.660951528397887\\
3.4	0.637172001131794\\
3.5	0.613947430401711\\
3.6	0.591361522566484\\
3.7	0.569477328049262\\
3.8	0.548339853524949\\
3.9	0.527978675408174\\
4	0.50841043135756\\
4.1	0.489641115145812\\
4.2	0.471668136459647\\
4.3	0.45448213286675\\
4.4	0.438068538459695\\
4.5	0.422408924583671\\
4.6	0.407482134311357\\
4.7	0.393265235341484\\
4.8	0.379734316841374\\
4.9	0.366865155217651\\
5	0.354633772435446\\
5.1	0.34301690866624\\
5.2	0.331992428903232\\
5.3	0.321539680742968\\
5.4	0.311639817601256\\
5.5	0.302276097775987\\
5.6	0.293434164237566\\
5.7	0.28510230167359\\
5.8	0.277271654587714\\
5.9	0.269936371429939\\
6	0.263093613733907\\
6.1	0.256743337543527\\
6.2	0.250887724591589\\
6.3	0.245530130897475\\
6.4	0.24067346128851\\
6.5	0.23631800372004\\
6.6	0.232458974991404\\
6.7	0.229084278523385\\
6.8	0.22617311220987\\
6.9	0.223695938824872\\
7	0.221615927693914\\
7.1	0.219891485433667\\
7.2	0.218479188289023\\
7.3	0.217336444674242\\
7.4	0.216423470927182\\
7.5	0.215704468254284\\
7.6	0.215148105201233\\
7.7	0.214727502143314\\
7.8	0.214419914969601\\
7.9	0.214206272471263\\
8	0.214070670729036\\
8.1	0.213999884815512\\
8.2	0.213982927859763\\
8.3	0.214010668607149\\
8.4	0.214075507966537\\
8.5	0.214171109752438\\
8.6	0.214292178720063\\
8.7	0.214434278615305\\
8.8	0.214593683439922\\
8.9	0.214767255963054\\
9	0.214952348425306\\
9.1	0.215146721249934\\
9.2	0.21534847634193\\
9.3	0.215556002205081\\
9.4	0.215767928643792\\
9.5	0.215983089253423\\
9.6	0.216200490255198\\
9.7	0.21641928451429\\
9.8	0.216638749805518\\
9.9	0.216858270571399\\
10	0.217077322561328\\
10.1	0.217295459855726\\
10.2	0.217512303871218\\
10.3	0.217727534016865\\
10.4	0.21794087973103\\
10.5	0.218152113676495\\
10.6	0.218361045910375\\
10.7	0.21856751887693\\
10.8	0.218771403097144\\
10.9	0.218972593449952\\
11	0.219171005957273\\
};
\addplot [color=mycolor2, mark=*, mark size=1pt, mark options={solid, mycolor2}, forget plot]
  table[row sep=crcr]{%
1.5	0.997307016602247\\
1.6	0.992906948917821\\
1.7	0.984215067481336\\
1.8	0.969440460151019\\
1.9	0.955374654420165\\
2	0.942678817915937\\
2.1	0.926791892941925\\
2.2	0.889753997835139\\
2.3	0.84073103175898\\
2.4	0.780734813546684\\
2.5	0.711999358492465\\
2.6	0.650946430151872\\
2.7	0.621049318819212\\
2.8	0.592089903638785\\
2.9	0.564047287602398\\
3	0.537023192596155\\
3.1	0.511098519349921\\
3.2	0.486328486735164\\
3.3	0.462744427090537\\
3.4	0.440356955121159\\
3.5	0.419159481011998\\
3.6	0.399131675191852\\
3.7	0.380242688286623\\
3.8	0.362454028257861\\
3.9	0.345722058707925\\
4	0.330000123736333\\
4.1	0.315240332163514\\
4.2	0.301395051709672\\
4.3	0.288418175296642\\
4.4	0.276266229894012\\
4.5	0.264899405386554\\
4.6	0.254282587642753\\
4.7	0.244386484358102\\
4.8	0.235188925815672\\
4.9	0.226676381987148\\
5	0.218845607993173\\
5.1	0.211705005669779\\
5.2	0.205274621187171\\
5.3	0.199582667362402\\
5.4	0.19465574186055\\
5.5	0.190501597711364\\
5.6	0.187089485983294\\
5.7	0.184339996507644\\
5.8	0.182133412481448\\
5.9	0.180331684339017\\
6	0.17880020563675\\
6.1	0.177420498100721\\
6.2	0.176093971582099\\
6.3	0.174740916580348\\
6.4	0.173298303028097\\
6.5	0.171718164524536\\
6.6	0.169966948026353\\
6.7	0.168025400931612\\
6.8	0.165888237685434\\
6.9	0.163562920690497\\
7	0.161067315076872\\
7.1	0.158426499512901\\
7.2	0.155669348638059\\
7.3	0.152825508059898\\
7.4	0.149923143799728\\
7.5	0.146987553780825\\
7.6	0.144040519942181\\
7.7	0.141100192445271\\
7.8	0.13818130023441\\
7.9	0.135295527281748\\
8	0.132451946744169\\
8.1	0.129657449423682\\
8.2	0.126917134034377\\
8.3	0.124234646300077\\
8.4	0.121612464987768\\
8.5	0.119052138589778\\
8.6	0.116554478716641\\
8.7	0.11411971681788\\
8.8	0.111747630505217\\
8.9	0.109437645025343\\
9	0.107188914594051\\
9.1	0.105000387497266\\
9.2	0.102870858146984\\
9.3	0.100799008669419\\
9.4	0.0987834420968172\\
9.5	0.0968227088222853\\
9.6	0.0949153276448644\\
9.7	0.0930598024660949\\
9.8	0.091254635487148\\
9.9	0.0894983375866363\\
10	0.0877894364247291\\
10.1	0.0861264827121117\\
10.2	0.0845080549969593\\
10.3	0.0829327632549062\\
10.4	0.0813992515124516\\
10.5	0.0799061996904889\\
10.6	0.0784523248194889\\
10.7	0.0770363817495289\\
10.8	0.075657163455482\\
10.9	0.074313501019146\\
11	0.0730042633550609\\
};
\addplot [color=mycolor3, mark=*, mark size=1pt, mark options={solid, mycolor3}, forget plot]
  table[row sep=crcr]{%
1.5	0.996007084097397\\
1.6	0.98943578612806\\
1.7	0.976209112057807\\
1.8	0.965796029465182\\
1.9	0.946154001506261\\
2	0.913991028665384\\
2.1	0.871715102323542\\
2.2	0.819488823071521\\
2.3	0.75875205983285\\
2.4	0.711751582722325\\
2.5	0.680870024934973\\
2.6	0.637647989116025\\
2.7	0.562918784034135\\
2.8	0.491011826229934\\
2.9	0.425112640926269\\
3	0.367136451629903\\
3.1	0.318079862912291\\
3.2	0.278407725853666\\
3.3	0.248118413516085\\
3.4	0.226050096165304\\
3.5	0.209836466646834\\
3.6	0.197332827030883\\
3.7	0.18729389442269\\
3.8	0.17911705515312\\
3.9	0.172525911666892\\
4	0.167372617239815\\
4.1	0.163516209896821\\
4.2	0.160769009217369\\
4.3	0.158907632900266\\
4.4	0.157710650236561\\
4.5	0.156985195747141\\
4.6	0.156574094765944\\
4.7	0.15635133871214\\
4.8	0.156213696241081\\
4.9	0.156072243010168\\
5	0.155845062739662\\
5.1	0.155451590961523\\
5.2	0.154809419370123\\
5.3	0.153835381747526\\
5.4	0.15245351426921\\
5.5	0.150610861881599\\
5.6	0.1482959935939\\
5.7	0.145548227368338\\
5.8	0.142448512626246\\
5.9	0.139096864248966\\
6	0.135590201744041\\
6.1	0.132009562015565\\
6.2	0.128416630442117\\
6.3	0.124855556265362\\
6.4	0.121356563760346\\
6.5	0.117939541411764\\
6.6	0.114616973906715\\
6.7	0.111396141560685\\
6.8	0.108280704828647\\
6.9	0.105271824851349\\
7	0.102368950300349\\
7.1	0.0995703693003131\\
7.2	0.0968735972657261\\
7.3	0.0942756501948803\\
7.4	0.0917732377350428\\
7.5	0.0893628997498611\\
7.6	0.0870411028526403\\
7.7	0.0848043083923848\\
7.8	0.0826490199635908\\
7.9	0.0805718161537993\\
8	0.0785693726053846\\
8.1	0.0766384763214792\\
8.2	0.074776034336588\\
8.3	0.0729790782964804\\
8.4	0.0712447660787695\\
8.5	0.0695703812869467\\
8.6	0.0679533312332028\\
8.7	0.0663911438660041\\
8.8	0.0648814639808681\\
8.9	0.0634220489656364\\
9	0.062010764266593\\
9.1	0.0606455787131817\\
9.2	0.0593245598025694\\
9.3	0.0580458690178206\\
9.4	0.0568077572327036\\
9.5	0.0556085602404888\\
9.6	0.054446694432271\\
9.7	0.0533206526414232\\
9.8	0.0522290001640849\\
9.9	0.0511703709605636\\
10	0.050143464038794\\
10.1	0.0491470400182584\\
10.2	0.0481799178707747\\
10.3	0.0472409718331509\\
10.4	0.0463291284857307\\
10.5	0.0454433639902231\\
10.6	0.0445827014798304\\
10.7	0.0437462085945039\\
10.8	0.0429329951541206\\
10.9	0.0421422109624376\\
11	0.0413730437348308\\
};
\end{axis}
\end{tikzpicture}%
  \end{subfigure}
  \caption{Comparison of the SDP embedding and the Diffusion Maps for the Wine dataset after a standardisation of the data points. Several different values of the bandwidth $\sigma$ are used.  In the embedding plots, each color refers a different class. At the m left, the non-zero eigenvalues of $B_\star/\Tr(B_\star)$ are plotted, whereas, at the bottom right, only the three largest eigenvalues of $\bar{\mathbf{A}}$ (see~\eqref{eq:DiscreteA}) are displayed. \rev{On the RHS, the diffusion embedding is displayed for normalized eigenvectors.}  \label{Fig:Wine}}
  \end{figure}
\subsection{Classification example} 
\begin{figure}[h]
  \centering
  \begin{minipage}{0.45\textwidth}
  \centering\textbf{SDP Embedding}\\
  \includegraphics[scale = 0.4]{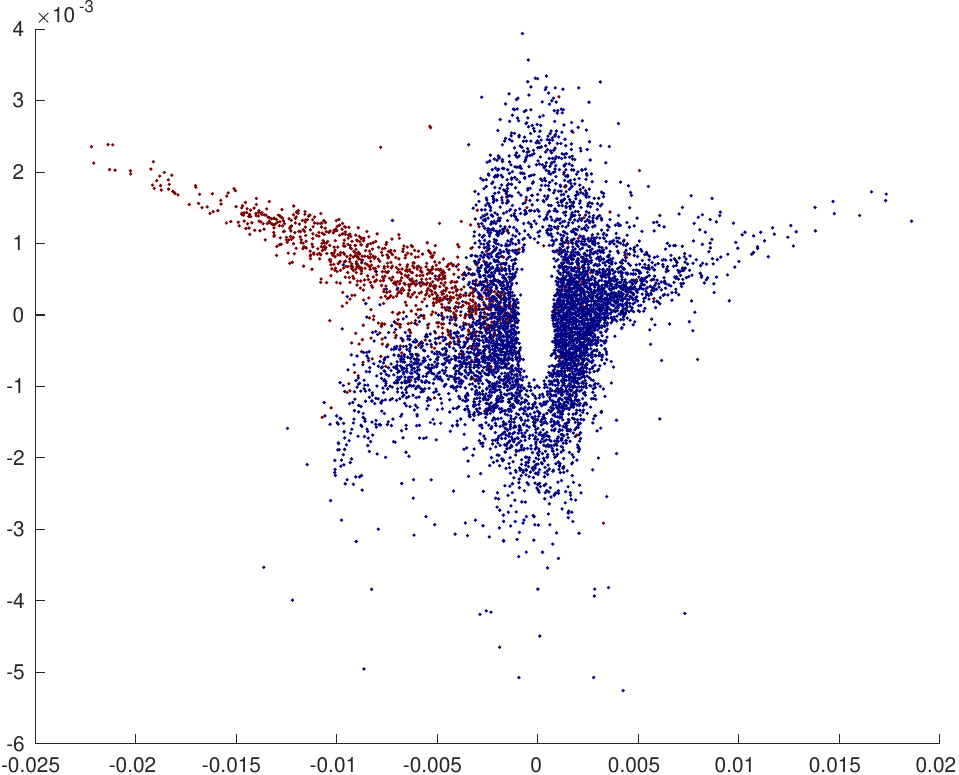}
  \end{minipage}\hfill
  \begin{minipage}{0.45\textwidth}
    \centering\textbf{Diffusion Embedding}\\
    \includegraphics[scale = 0.4]{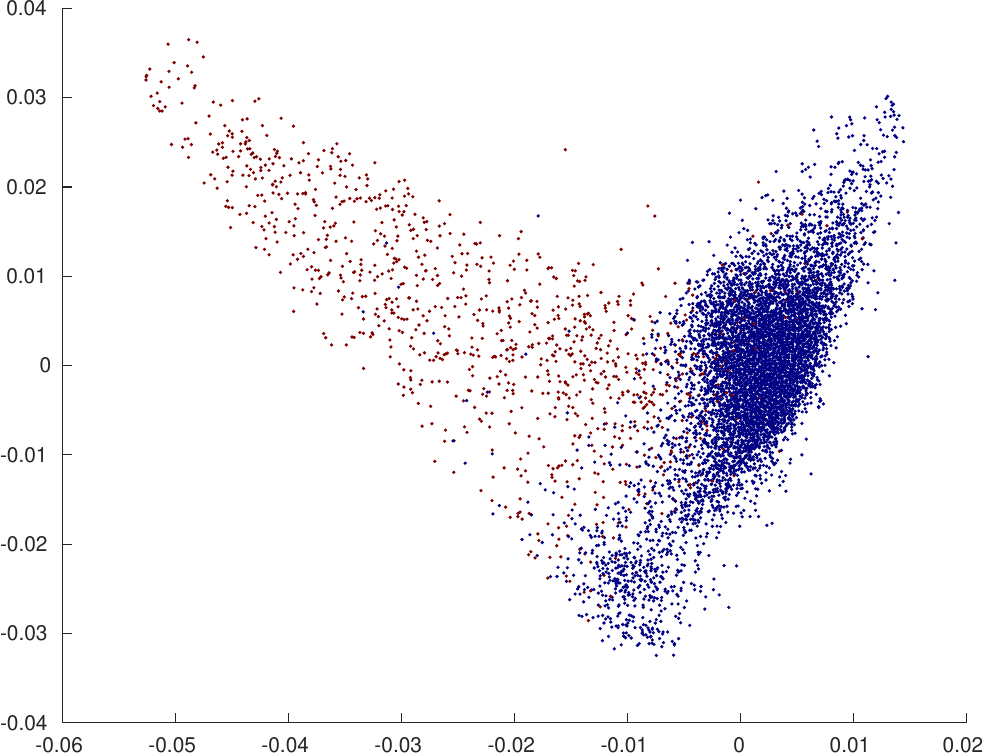}
    \end{minipage}
  \caption{Embedding of the training set of standardized HTRU2 dataset for $\sigma = 10$. Pulsars are in red and non-pulsars in blue. Left: the SDP embedding dimension is the number of numerically significant eigenvalues of $\mathbf{B}_\star/\Tr(\mathbf{B}_\star)$, namely $2$ in this case. Right: Diffusion embedding with $2$ normalized eigenvectors and $\sigma = 10$ .  \label{fig:pulsar}}
  \end{figure}

A larger scale example is given in Figure~\ref{fig:pulsar} which displays the embedding of the astrophysics HTRU2 dataset. 
 As in the other examples, this SDP embedding has a annulus shape which can be intuitively interpreted as follows: the points on the same radius have a similar `centrality' in the dataset. In Figure~\ref{fig:pulsar}, the minority class (pulsars, in red) can be seen as a spike on top of the annulus.
 A $k$-nearest neighbours classifier is trained $3$ times on the SDP embedding of $70$ percent of the dataset (uniform sample) with $k=5$. The out-of-sample formula~\eqref{eq:Out-of-sample2} is used to predict on the test set, i.e., the remaining $30$ percent of the dataset. The results are reported in Table~\ref{tab:HTRU2}, where we observe that the Diffusion and SDP embeddings yield almost the same precision and recall for this classification task.
 \begin{table}[h]
\begin{center}
  \resizebox{\textwidth}{!}{
  \begin{tabular}{l l l l l l l l}
      \toprule
      \multicolumn{4}{c}{SDP embedding} & \multicolumn{4}{c}{Diffusion embedding}\\
      \cmidrule(lr){1-4}\cmidrule(lr){5-8}
      \multicolumn{2}{c}{$\sigma = 10$} & \multicolumn{2}{c}{$\sigma = 5$} & \multicolumn{2}{c}{$\sigma = 10$} & \multicolumn{2}{c}{$\sigma = 5 $}\\
      \cmidrule(lr){1-2}       \cmidrule(lr){3-4} \cmidrule(lr){5-6} \cmidrule(lr){7-8}
      precision & recall & precision & recall & precision & recall & precision & recall\\
      0.90(0.01) & 0.76(0.02) & 0.91(0.01)  & 0.79(0.01) & 0.90(0.01)  & 0.78(0.02) & 0.91(0.01) & 0.79(0.01)\\
      \bottomrule
  \end{tabular}
  }
\end{center}
\caption{Classification results for the SDP and diffusion embedding (with two components) of the HTRU2 dataset. The positive class is here `pulsars' (minority).  The standard deviation over $3$ runs is given in parenthesis. \label{tab:HTRU2}}
\end{table}
\subsection{Clustering example}
\rev{
  SDP embedding is also applied to a commonly used artificial two-moons dataset displayed in Figure~\ref{fig:twomoons}. After embedding, the dataset is clustered thanks to $k$-means. In Figure~\ref{fig:clustering_performance}, we display the Normalized Mutual Information (NMI) between the cluster label vector and the ground truth for several values of the Gaussian kernel's bandwidth. Compared with the diffusion embedding\footnote{The eigenvalue decomposition sometimes fails to converge when $\sigma$ is small.}, we observe that the clustering on the SDP embedding has a higher performance with a lower variance. In the same figure, a similar procedure is applied on a post-processed embedding obtained by projecting all embedded points on a unit circle; see e.g.~\cite{QinRohe}. As expected, the performance of diffusion embedding gets then closer to the SDP embedding performance although the latter has still a smaller variance. 
}
\begin{figure}[h]
  \centering
  \figureheight = 0.2\textwidth
  \figurewidth = 0.4\textwidth
  \begin{minipage}{0.5\textwidth}
    \centering\textbf{\small Original data }\\
    \input{Figures/twomoons.tikz}
  \end{minipage}\hfill
  \begin{minipage}{0.5\textwidth}
    \centering\textbf{\small SDP embedding }\\
    \input{Figures/twomoonsSDP.tikz}
  \end{minipage}\hfill
  \caption{\rev{Illustration of the two-moons dataset with $200$ points in each cluster. Raw data (left). SDP embedding of the standardized data with the Gaussian kernel for $\sigma = 0.1$ (right).} \label{fig:twomoons} }
\end{figure}
\begin{figure}[h]
  \centering
  \begin{minipage}{0.5\textwidth}
  \centering\textbf{\small Clustering on embedding}\\
  \figureheight = 0.4\textwidth
  \figurewidth = 0.8\textwidth
%
%
\begin{tikzpicture}

\begin{axis}[%
width=\figurewidth,
height=\figureheight,
at={(1.011in,0.642in)},
scale only axis,
xlabel = {$\sigma$},
xmin=0,
xmax=1,
ymin=0,
ymax=1,
axis background/.style={fill=white},
legend style={legend cell align=left, align=left, draw=white!15!black}
]
\addplot [color=black]
 plot [error bars/.cd, y dir = both, y explicit]
 table[row sep=crcr, y error plus index=2, y error minus index=3]{%
0.05	0.355432350541462	0.0114172300400009	0.0114172300400009\\
0.1	0.355432350541462	0.0114172300400009	0.0114172300400009\\
0.15	0.355432350541462	0.0114172300400009	0.0114172300400009\\
0.2	0.355432350541462	0.0114172300400009	0.0114172300400009\\
0.25	0.355432350541462	0.0114172300400009	0.0114172300400009\\
0.3	0.355432350541462	0.0114172300400009	0.0114172300400009\\
0.35	0.355432350541462	0.0114172300400009	0.0114172300400009\\
0.4	0.355432350541462	0.0114172300400009	0.0114172300400009\\
0.45	0.355432350541462	0.0114172300400009	0.0114172300400009\\
0.5	0.355432350541462	0.0114172300400009	0.0114172300400009\\
0.55	0.355432350541462	0.0114172300400009	0.0114172300400009\\
0.6	0.355432350541462	0.0114172300400009	0.0114172300400009\\
0.65	0.355432350541462	0.0114172300400009	0.0114172300400009\\
0.7	0.355432350541462	0.0114172300400009	0.0114172300400009\\
0.75	0.355432350541462	0.0114172300400009	0.0114172300400009\\
0.8	0.355432350541462	0.0114172300400009	0.0114172300400009\\
0.85	0.355432350541462	0.0114172300400009	0.0114172300400009\\
0.9	0.355432350541462	0.0114172300400009	0.0114172300400009\\
0.95	0.355432350541462	0.0114172300400009	0.0114172300400009\\
1	0.355432350541462	0.0114172300400009	0.0114172300400009\\
};
\addlegendentry{\footnotesize NMI kmeans}

\addplot [color=blue, mark=o, mark options={solid, blue}]
 plot [error bars/.cd, y dir = both, y explicit]
 table[row sep=crcr, y error plus index=2, y error minus index=3]{%
0.05	0.046432008104814	0.0382163449604739	0.0382163449604739\\
0.1	0.957197654838908	0.0343572302876194	0.0343572302876194\\
0.15	0.91117303832729	0.144850967233726	0.144850967233726\\
0.2	0.809582604349096	0.223716927842793	0.223716927842793\\
0.25	0.734418498264921	0.223980794357623	0.223980794357623\\
0.3	0.672629264274321	0.219740116142731	0.219740116142731\\
0.35	0.5474623938608	0.136395628660129	0.136395628660129\\
0.4	0.490017176074386	0.0724998462070194	0.0724998462070194\\
0.45	0.446134815865321	0.0392119851320656	0.0392119851320656\\
0.5	0.427989901173205	0.0279498000698364	0.0279498000698364\\
0.55	0.408795112964163	0.0258763831044252	0.0258763831044252\\
0.6	0.401048148659463	0.0256416918332774	0.0256416918332774\\
0.65	0.395963427631874	0.026284253252487	0.026284253252487\\
0.7	0.391363416939119	0.0241935967585841	0.0241935967585841\\
0.75	0.389466125838816	0.0237473117345404	0.0237473117345404\\
0.8	0.384337840722041	0.0191220408344743	0.0191220408344743\\
0.85	0.383765497469846	0.0212130084488161	0.0212130084488161\\
0.9	0.380065887637544	0.0208183296141971	0.0208183296141971\\
0.95	0.378173089115287	0.0195047241182366	0.0195047241182366\\
1	0.376368681591752	0.0198531521043574	0.0198531521043574\\
};
\addlegendentry{\footnotesize NMI SDP}

\addplot [color=red, mark=square, mark options={solid, red}]
 plot [error bars/.cd, y dir = both, y explicit]
 table[row sep=crcr, y error plus index=2, y error minus index=3]{%
0.05	0.015773278171576	0	0\\
0.1	0.590921963074662	0.494125694099906	0.494125694099906\\
0.15	0.42982323360866	0.483504314532028	0.483504314532028\\
0.2	0.610887681728574	0.440274227493761	0.440274227493761\\
0.25	0.278040652339336	0.29186699279682	0.29186699279682\\
0.3	0.547967871163347	0.415818212977073	0.415818212977073\\
0.35	0.276159281370274	0.222181989070298	0.222181989070298\\
0.4	0.469606413130355	0.243925009229462	0.243925009229462\\
0.45	0.377134589950457	0.214743194330335	0.214743194330335\\
0.5	0.420339567511854	0.207602519047824	0.207602519047824\\
0.55	0.344666968146268	0.19815462044504	0.19815462044504\\
0.6	0.222549303875316	0.170624166131654	0.170624166131654\\
0.65	0.250934565124715	0.179144070009898	0.179144070009898\\
0.7	0.370904906930491	0.12726230212565	0.12726230212565\\
0.75	0.330442847583155	0.142409961106745	0.142409961106745\\
0.8	0.386703318710084	0.0827522283961808	0.0827522283961808\\
0.85	0.401815606238556	0.0142750520347631	0.0142750520347631\\
0.9	0.394887902259317	0.0160258731022029	0.0160258731022029\\
0.95	0.394162812668709	0.0147677225196379	0.0147677225196379\\
1	0.387950354604084	0.0166653018115305	0.0166653018115305\\
};
\addlegendentry{\footnotesize NMI DM}

\end{axis}
\end{tikzpicture}%
  \figureheight = 0.2\textwidth
  \centering\textbf{\small Spectrum of $\mathbf{B}_\star/\Tr(\mathbf{B}_\star)$}\\
%
%
\definecolor{mycolor1}{rgb}{0.00000,0.44700,0.74100}%
\definecolor{mycolor2}{rgb}{0.85000,0.32500,0.09800}%
\definecolor{mycolor3}{rgb}{0.92900,0.69400,0.12500}%
\definecolor{mycolor4}{rgb}{0.49400,0.18400,0.55600}%
\begin{tikzpicture}

\begin{axis}[%
width=\figurewidth,
height=\figureheight,
at={(1.011in,0.642in)},
scale only axis,
xmin=0,
xmax=1,
ymin=0,
ymax=1,
xlabel = {$\sigma$},
axis background/.style={fill=white},
legend style={legend cell align=left, align=left, draw=white!15!black}
]
\addplot [color=mycolor1, mark size=1pt, mark=*, mark options={solid, mycolor1}]
  table[row sep=crcr]{%
0.05	0.136298813342149\\
0.1	0.985157212203279\\
0.15	0.976205207646594\\
0.2	0.841208432838684\\
0.25	0.731075614028974\\
0.3	0.696544053035181\\
0.35	0.679538361459447\\
0.4	0.672051073523674\\
0.45	0.67145618989876\\
0.5	0.675668267704874\\
0.55	0.68317193930884\\
0.6	0.693087953193807\\
0.65	0.704881742718947\\
0.7	0.718165392243112\\
0.75	0.732617884759328\\
0.8	0.747917244559839\\
0.85	0.763340582156343\\
0.9	0.778399478366726\\
0.95	0.793844106132185\\
1	0.809370142045361\\
};

\addplot [color=mycolor2, mark size=1pt, mark=*, mark options={solid, mycolor2}]
  table[row sep=crcr]{%
0.05	0.108054814823416\\
0.1	0.00683109645669601\\
0.15	0.0237899476322117\\
0.2	0.148939893185632\\
0.25	0.15660007038798\\
0.3	0.196670566649177\\
0.35	0.202991505167006\\
0.4	0.201024378754626\\
0.45	0.200980803588702\\
0.5	0.204976714123368\\
0.55	0.211563065154399\\
0.6	0.218919269126678\\
0.65	0.225878204570499\\
0.7	0.231822658483348\\
0.75	0.23643142089637\\
0.8	0.239532261504067\\
0.85	0.236659417843654\\
0.9	0.221600521633274\\
0.95	0.206155893867815\\
1	0.190629857954639\\
};

\addplot [color=mycolor3, mark size=1pt, mark=*, mark options={solid, mycolor3}]
  table[row sep=crcr]{%
0.05	0.0882119044934605\\
0.1	0.00659735669036575\\
0.15	4.84444060377315e-06\\
0.2	0.00985167397568459\\
0.25	0.112324315583046\\
0.3	0.106785380315642\\
0.35	0.117470133373547\\
0.4	0.126924547721701\\
0.45	0.127563006512537\\
0.5	0.119355018171758\\
0.55	0.105264995536761\\
0.6	0.0879927776795152\\
0.65	0.0692400527105539\\
0.7	0.0500119492735401\\
0.75	0.0309506943443018\\
0.8	0.0125504939360939\\
0.85	3.0204536149084e-15\\
0.9	1.43871967623911e-16\\
0.95	4.71139358806671e-17\\
1	2.29308189134978e-17\\
};

\addplot [color=mycolor4, mark size=1pt, mark=*, mark options={solid, mycolor4}]
  table[row sep=crcr]{%
0.05	0.0816207032061886\\
0.1	0.000695660231465667\\
0.15	2.80590426574709e-10\\
0.2	8.74404835662742e-32\\
0.25	5.54333301533137e-32\\
0.3	4.37619559439805e-32\\
0.35	2.51528058881932e-32\\
0.4	3.32850325666884e-32\\
0.45	3.24022890370987e-32\\
0.5	2.34203073248336e-32\\
0.55	2.78336204706011e-31\\
0.6	2.19116014780799e-32\\
0.65	2.20210504643446e-32\\
0.7	2.50352644246361e-32\\
0.75	3.68555646405841e-32\\
0.8	2.83327055796526e-32\\
0.85	2.75324635870068e-32\\
0.9	3.34270001959447e-32\\
0.95	1.86136581815556e-32\\
1	2.74957072715398e-32\\
};

\end{axis}
\end{tikzpicture}%
  \end{minipage}\hfill
  \begin{minipage}{0.5\textwidth}
    \centering\textbf{\small Clustering on projected Embedding}\\
    \figureheight = 0.4\textwidth
    \figurewidth = 0.8\textwidth
%
%
\begin{tikzpicture}

\begin{axis}[%
width=\figurewidth,
height=\figureheight,
at={(1.011in,0.642in)},
scale only axis,
xlabel = {$\sigma$},
xmin=0,
xmax=1,
ymin=0,
ymax=1,
axis background/.style={fill=white},
legend style={legend cell align=left, align=left, draw=white!15!black}
]
\addplot [color=black]
 plot [error bars/.cd, y dir = both, y explicit]
 table[row sep=crcr, y error plus index=2, y error minus index=3]{%
0.05	0.355432350541462	0.0114172300400009	0.0114172300400009\\
0.1	0.355432350541462	0.0114172300400009	0.0114172300400009\\
0.15	0.355432350541462	0.0114172300400009	0.0114172300400009\\
0.2	0.355432350541462	0.0114172300400009	0.0114172300400009\\
0.25	0.355432350541462	0.0114172300400009	0.0114172300400009\\
0.3	0.355432350541462	0.0114172300400009	0.0114172300400009\\
0.35	0.355432350541462	0.0114172300400009	0.0114172300400009\\
0.4	0.355432350541462	0.0114172300400009	0.0114172300400009\\
0.45	0.355432350541462	0.0114172300400009	0.0114172300400009\\
0.5	0.355432350541462	0.0114172300400009	0.0114172300400009\\
0.55	0.355432350541462	0.0114172300400009	0.0114172300400009\\
0.6	0.355432350541462	0.0114172300400009	0.0114172300400009\\
0.65	0.355432350541462	0.0114172300400009	0.0114172300400009\\
0.7	0.355432350541462	0.0114172300400009	0.0114172300400009\\
0.75	0.355432350541462	0.0114172300400009	0.0114172300400009\\
0.8	0.355432350541462	0.0114172300400009	0.0114172300400009\\
0.85	0.355432350541462	0.0114172300400009	0.0114172300400009\\
0.9	0.355432350541462	0.0114172300400009	0.0114172300400009\\
0.95	0.355432350541462	0.0114172300400009	0.0114172300400009\\
1	0.355432350541462	0.0114172300400009	0.0114172300400009\\
};
\addlegendentry{\footnotesize NMI kmeans}

\addplot [color=blue, mark=o, mark options={solid, blue}]
 plot [error bars/.cd, y dir = both, y explicit]
 table[row sep=crcr, y error plus index=2, y error minus index=3]{%
0.05	0.033807385223082	0.0317580027436102	0.0317580027436102\\
0.1	0.957197654838908	0.0343572302876194	0.0343572302876194\\
0.15	0.909997274550402	0.143957085732976	0.143957085732976\\
0.2	0.795144840052501	0.21720614835765	0.21720614835765\\
0.25	0.74426163810876	0.21987708127379	0.21987708127379\\
0.3	0.689316144666546	0.215579377400781	0.215579377400781\\
0.35	0.574110155766427	0.136045799025743	0.136045799025743\\
0.4	0.502365581356902	0.0739435376259482	0.0739435376259482\\
0.45	0.452439773197694	0.0384638725806322	0.0384638725806322\\
0.5	0.429983658883532	0.0265942969824545	0.0265942969824545\\
0.55	0.413312655888356	0.0243468958783592	0.0243468958783592\\
0.6	0.406625396046426	0.0212491789014	0.0212491789014\\
0.65	0.399098636032578	0.0259520663283014	0.0259520663283014\\
0.7	0.39581524921578	0.0245714086646096	0.0245714086646096\\
0.75	0.391967229938879	0.0240677971713336	0.0240677971713336\\
0.8	0.386844197839503	0.0207098264395343	0.0207098264395343\\
0.85	0.385572970815117	0.0195016484098216	0.0195016484098216\\
0.9	0.382475030503129	0.0194440352837964	0.0194440352837964\\
0.95	0.381213571398988	0.0187668217223433	0.0187668217223433\\
1	0.378806550099607	0.0202095572197988	0.0202095572197988\\
};
\addlegendentry{\footnotesize NMI SDP+proj}

\addplot [color=red, mark=square, mark options={solid, red}]
 plot [error bars/.cd, y dir = both, y explicit]
 table[row sep=crcr, y error plus index=2, y error minus index=3]{%
0.05	0.015773278171576	0	0\\
0.1	0.778316515160903	0.384882782780325	0.384882782780325\\
0.15	0.77809992177429	0.409230842376972	0.409230842376972\\
0.2	0.674710301441193	0.426866085933634	0.426866085933634\\
0.25	0.810591943223703	0.276299982691529	0.276299982691529\\
0.3	0.720880406475339	0.275975058114037	0.275975058114037\\
0.35	0.575725198211973	0.311087576145212	0.311087576145212\\
0.4	0.505741368468059	0.337875998319592	0.337875998319592\\
0.45	0.503525184248843	0.260821158765997	0.260821158765997\\
0.5	0.393442211161654	0.22444372845484	0.22444372845484\\
0.55	0.36165930401097	0.201395880842087	0.201395880842087\\
0.6	0.29412094337043	0.198977825947165	0.198977825947165\\
0.65	0.391753025814693	0.147452802428867	0.147452802428867\\
0.7	0.273806671049521	0.158486052525337	0.158486052525337\\
0.75	0.325780768763944	0.155942746981417	0.155942746981417\\
0.8	0.317859479738566	0.152963712097003	0.152963712097003\\
0.85	0.375202567613432	0.103145491955049	0.103145491955049\\
0.9	0.361829613058189	0.127555585571778	0.127555585571778\\
0.95	0.393529893627299	0.0157384936850168	0.0157384936850168\\
1	0.391123785224488	0.0186344380931787	0.0186344380931787\\
};
\addlegendentry{\footnotesize NMI DM+proj}

\end{axis}
\end{tikzpicture}%
    \centering\textbf{\small Spectrum of $\bar{\mathbf{A}}$}\\
    \figureheight = 0.2\textwidth
%
%
\definecolor{mycolor1}{rgb}{0.00000,0.44700,0.74100}%
\definecolor{mycolor2}{rgb}{0.85000,0.32500,0.09800}%
\definecolor{mycolor3}{rgb}{0.92900,0.69400,0.12500}%
\definecolor{mycolor4}{rgb}{0.49400,0.18400,0.55600}%
\begin{tikzpicture}

\begin{axis}[%
width=\figurewidth,
height=\figureheight,
at={(1.011in,0.642in)},
scale only axis,
xlabel={$\sigma$},
unbounded coords=jump,
xmin=0.1,
xmax=1,
ymin=0.5,
ymax=1,
axis background/.style={fill=white},
legend style={legend cell align=left, align=left, draw=white!15!black}
]
\addplot [color=mycolor1, mark size=1pt, mark=*, mark options={solid, mycolor1}]
  table[row sep=crcr]{%
0.05	nan\\
0.1	0.999999391698469\\
0.15	0.999932188541798\\
0.2	0.999415025267962\\
0.25	0.997929040727315\\
0.3	0.995373102444973\\
0.35	0.991891992980415\\
0.4	0.987503024853293\\
0.45	0.982146096949744\\
0.5	0.975779057686984\\
0.55	0.968393140630774\\
0.6	0.96000107706843\\
0.65	0.950625823226078\\
0.7	0.94029465445193\\
0.75	0.929036599797351\\
0.8	0.916882184462181\\
0.85	0.903865104477624\\
0.9	0.890024730819009\\
0.95	0.875408008229528\\
1	0.860069910066966\\
};

\addplot [color=mycolor2, mark size=1pt, mark=*, mark options={solid, mycolor2}]
  table[row sep=crcr]{%
0.05	nan\\
0.1	0.999696897631623\\
0.15	0.998359125165173\\
0.2	0.995939402801231\\
0.25	0.991702515485207\\
0.3	0.98457830242055\\
0.35	0.974226096181291\\
0.4	0.960817095284076\\
0.45	0.944391655136474\\
0.5	0.925164470199371\\
0.55	0.90339664689587\\
0.6	0.879374449068874\\
0.65	0.853440444382131\\
0.7	0.825977882726541\\
0.75	0.797374900649006\\
0.8	0.767997927659587\\
0.85	0.738180360247819\\
0.9	0.708221143961132\\
0.95	0.678386626121053\\
1	0.64891183212578\\
};

\addplot [color=mycolor3, mark size=1pt, mark=*, mark options={solid, mycolor3}]
  table[row sep=crcr]{%
0.05	nan\\
0.1	0.999695601753321\\
0.15	0.998267406675155\\
0.2	0.995542256118285\\
0.25	0.990922853872129\\
0.3	0.983587711967758\\
0.35	0.972675068566217\\
0.4	0.957800084568779\\
0.45	0.939015589127466\\
0.5	0.916253521592662\\
0.55	0.88962253712314\\
0.6	0.859518426927576\\
0.65	0.826516699783109\\
0.7	0.791255082455187\\
0.75	0.754361220063415\\
0.8	0.716419088385\\
0.85	0.677957736715473\\
0.9	0.639449232676917\\
0.95	0.601308663817733\\
1	0.563893884603461\\
};

\addplot [color=mycolor4, mark size=1pt, mark=*, mark options={solid, mycolor4}]
  table[row sep=crcr]{%
0.05	nan\\
0.1	0.99882670804497\\
0.15	0.993449602263227\\
0.2	0.985357718136186\\
0.25	0.974210636035643\\
0.3	0.959465560508284\\
0.35	0.940868840443262\\
0.4	0.918608790698895\\
0.45	0.893064051070515\\
0.5	0.864616138123337\\
0.55	0.833646542222768\\
0.6	0.800565154036679\\
0.65	0.765814802670842\\
0.7	0.729857439690762\\
0.75	0.69315468865128\\
0.8	0.656149349356336\\
0.85	0.619249861690723\\
0.9	0.582818461086133\\
0.95	0.547163644264908\\
1	0.512537193062334\\
};

\end{axis}
\end{tikzpicture}%
    \end{minipage}
  \caption{\rev{Clustering performance on the two-moons dataset. Top row: Normalized Mutual Information (NMI, the larger the better) between the ground truth and the clustering obtained on the embeddings, vs the kernel bandwidth $\sigma$. On the LHS, $k$-means clustering with $k=2$ is run once on a 2-dimensional SDP embedding (blue circles), a 2-dimensional DM embedding (red squares) and on the raw data (black), as a baseline. On the RHS, a similar procedure is performed excepted that the embedding is post-processed by projecting all points on a circle before applying $k$-means.  The markers denote the NMIs averaged over 10 runs, whereas the error bars are standard deviations. Bottom row: we display, as a function  of $\sigma$, the four leading eigenvalues of $\mathbf{B}_\star/\Tr(\mathbf{B}_\star)$ on the left and the four leading eigenvalues of the matrix $\bar{\mathbf{A}}$, defined in~\eqref{eq:DiscreteA}, on the right. Again, from the observation of the eigenvalues of $\mathbf{B}_\star/\Tr(\mathbf{B}_\star)$, we notice that the effective dimension of the SDP embedding is very low.} \label{fig:clustering_performance}}
  \end{figure}
\section{Related work} We now review a few related papers.
\paragraph{Optimization problem} The type of optimization problem in finite dimension solved in this paper has been studied in various works in different contexts. Several authors examined the connection between an SDP and the max-cut problem, $\mathbb{Z}_2$ or angular synchronization, as well as community detection~\cite{Burer2003,BoumalGeNPowMeth,MontanariPNAS,Chretienetal}. Also, in the context of the angular synchronization problem, the paper~\cite{Bandeira2016} investigates when an SDP relaxation has a global rank one solution which solves exactly an angular synchronization problem. The conclusions presented here can be viewed as a generalization of previous ideas applied to graph data~\cite{Burer2003,BoumalGeNPowMeth,Bandeira2016,MontanariPNAS,MasterThesis} in the context of kernel methods. Concerning the scalability of (structured) SDPs, \cite{scalableSDP} recently provided an efficient optimization strategy with theoretical guarantees for solving large SDPs with the help of sketching. The latter work might be a source of inspiration for scaling up the approach presented here.

\paragraph{Learning a positive semi-definite matrix} Another dimensionality reduction technique associated to an SDP is the so-called Maximum Variance Unfolding (MVU), also called Semidefinite Embedding, introduced in~\cite{WeinbergerICML}, and used for computer vision in~\cite{WeinbergerCVPR}. Although similar in spirit, the objective function of MVU differs from the objective considered in this paper. Furthermore, an asset of our approach is the out-of-sample formula.
\paragraph{Learning a kernel in a RKHS}
Learning the kernel is a topic of interest in the context of supervised learning and has been investigated e.g. in ~\cite{Micchelli2005,Micchelli2016} in the framework of Reproducing Kernel Hilbert Spaces. To the best of our knowledge, the variational problem in infinite dimensions defined in this paper has not yet been studied in the literature. Let us also mention that, in another context, kernels of nuclear operators have been studied for instance in~\cite{Vershik2013,Vershik2015} in connection with the virtual continuity.
\paragraph{State-of-the-art dimensionality reduction} Two leading dimensionality reduction methods are  t-distributed Stochastic Neighbor Embedding (t-SNE~\cite{vanDerMaaten2008}) and Uniform Manifold Approximation and Projection (UMAP~\cite{UMAP}). These methods are highly successful in practice and scalable. An advantage of our approach is its interpretability and the out-of-sample formula, while a major drawback is that SDP embedding does not achieve as good visualizations in two dimensions as t-SNE or UMAP.
\section{Conclusions}

In this work, we discussed a novel dimensionality reduction technique which is based on a semi-definite program. Two approaches were presented: with weak and strong assumptions about smoothness. An out-of-sample formula is also provided in both cases. It is observed numerically that the embedding is robust to the presence of outliers. 
Possible future research includes the empirical study of the learning performance of the finite dimensional feature maps defined by the out-of-sample extension formula given by the kernelized problem of Section~\ref{sec:Kernelized}. The numerical simulations reported here only considered the Gaussian kernel with weak smoothness assumptions. It would be instructive to analyse the role of the regularization parameter of the kernelized problem as well as the influence of the regularity parameter of the Sobolev kernel. 

\FloatBarrier
\appendix
\section{Useful elements of operator theory}
\subsection{Hilbert-Schmidt and Nuclear operators}
Let $(\mathcal{H}, \langle \cdot,\cdot\rangle)$ be a separable Hilbert space with orthonormal basis (ONB) $\{\phi_\ell\}_{\ell\geq 1}$. We say that $A:\mathcal{H}\to \mathcal{H}$ is a compact linear operator if for any bounded sequence $(f_\ell)_\ell\in\mathcal{H}$, the sequence  $(Af_\ell)_\ell\in\mathcal{H}$ admits a convergent subsequence. We denote the adjoint of $A$ by $A^*$. An operator $A$ is \emph{psd} if $\langle v, Av\rangle \geq 0$ for all $v\in \mathcal{H}$. The operator $A$ is Hilbert-Schmidt if $\sum_{\ell\geq 1} \langle A \phi_\ell ,A \phi_\ell\rangle <\infty$. The Hilbert-Schmidt operators form a Hilbert space for the inner product
$
\langle A, B\rangle_{HS} = \sum_{\ell\geq 1}\langle A \phi_\ell ,B \phi_\ell\rangle.
$
A Hilbert-Schmidt operator $A$ is nuclear (or trace class) if
$\sum_{\ell\geq 1}\langle \sqrt{A^* A} \phi_\ell , \phi\ell\rangle< \infty,
$
while the trace of this operator reads $\Tr(A) = \sum_{\ell\geq 1} \langle A \phi_\ell , \phi_\ell\rangle$. Finally, the trace class or nuclear norm of an operator is $\|A\|_{*} = \Tr(\sqrt{A^* A})$. Notice that these definitions are in fact independent of the choice of ONB for $\mathcal{H}$.
\subsection{Reproducing Kernel Hilbert Space}
Let $X$ be a set. A kernel is a symmetric function $k:X\times X\to\mathbb{R}$.  A kernel is positive semi-definite (\emph{psd}) if for all finite samples $\{x_1, \dots , x_m\}$ of points in $X$, the corresponding Gram matrix $[\mathbf{K}]_{ij} = k(x_i, x_j)$ which is positive semi-definite. Let $\phi(x):X\to \mathbb{R}$ be the function $k(x,\cdot)$. The Reproducing Kernel Hilbert Space (RKHS) $\mathcal{H}_k$ is given by the completion of ${\rm span}\{ \phi(x) \text{  s.t.  } x\in X\}$ endowed with the inner product $\langle \phi(x), \phi(y)\rangle = k(x, y)$.

\subsection{Generalized Mercer Theorem}
\begin{theorem}[Mercer's theorem for \emph{psd} symmetric nuclear operators, Thm 3.10 in~\cite{Steinwart2012}]\label{Thm:NuclearDecomposition}
 Let $X$ be a measurable space, $\mu$ be a measure on $X$, and $K :L^2(X,\mu)\to L^2(X,\mu)$ be a psd, symmetric and nuclear operator. We write $\{\lambda^{(\ell)}\}_{\ell\geq 1}$ for the at most countably many non-zero eigenvalues of $K$, where we included geometric multiplicities. Then there exists a measurable kernel $k$ on $X$ with separable RKHS $\mathcal{H}_k$ such that
 $
( K f)(\cdot) = \int_X k(\cdot, y)f(y)\rmd \mu(y),
 $ with $f \in L^2(X,\mu)$ and
 where the left-hand side of the equation is considered to be a $\mu$-equivalence class in $L^2(X,\mu)$. In addition, $k$ enjoys a Mercer representation, that is, there exists an ONB $\{\sqrt{\lambda^{(\ell)}}\phi^{(\ell)}\}_{\ell\geq 1}$ of $\mathcal{H}_k$ such that $\{\phi^{(\ell)}\}_{\ell\geq 1}$ is an orthonormal system in $L^2(X,\mu)$ consisting of the eigenfunctions corresponding to the eigenvalues $\{\lambda^{(\ell)}\}_{\ell\geq 1}$ of $K$ such that
\begin{equation*}
k(x,y) = \sum_{\ell\geq 1} \lambda^{(\ell)} \phi^{(\ell)}(x)\phi^{(\ell)}(y),  \text{ for all } x,y\in X.
\end{equation*}
Moreover, the integral operator $K$ is defined pointwise, that is, $Kf(x) = \int_X k(x, y)f(y)\rmd \mu(y), $
 for all $ f\in L^2(X,\mu)$ and all $x\in X$.
\end{theorem}

\section{Dual certificate\label{sec:Dual}}
By using the duality theory of semi-definite programs, we can show that the optimality of a matrix $\mathbf{B}$ can be certified thanks to a so-called dual certificate.
Indeed, we can write the Lagrangian associated to \eqref{eq:SDP}, 
$
\mathcal{L}(\mathbf{B},\bm{y}) =  \Tr\left(\mathbf{\bar{A}} \mathbf{B}\right)   +  \bm{y}^\top \left(\diag(\mathbf{B})- \bm{d}\right).
$
The primal optimization problem is
$p_\star = \max_{\mathbf{B}\succeq 0}\min_{\bm{y}\in \mathbb{R}^n}\mathcal{L}(\mathbf{B},\bm{y})$. The dual problem, which is classically given by $d_\star = \min_{\bm{y}\in \mathbb{R}^n} \max_{\mathbf{B}\succeq 0}\mathcal{L}(\mathbf{B},\bm{y})$,  can be simplified as
\[
d_\star =\min_{\bm{y}\in \mathbb{R}^n}  \bm{d}^\top \bm{y}, \text{ subject to } \Diag(\bm{y}) -\mathbf{\bar{A}}\succeq 0.
\]
Since the matrix $\mathbf{B} = \Diag(\bm{d})\succ 0$ is strictly feasible for \eqref{eq:SDP} (Slater condition), then the optimal values of the dual and primal problems coincide, namely $d_\star =p_\star$.  The duality gap can be written as follows
\[
d_\star - p_\star = \bm{d}^\top \bm{y}_\star -\Tr\left(\mathbf{\bar{A}} \mathbf{B}_\star\right) = \Tr\Big((\Diag(\bm{y}_\star)-\mathbf{\bar{A}})\mathbf{B}_\star\Big).
\]
Since by feasibility $\mathbf{B}_\star\succeq 0$ and $\Diag(\bm{y}_\star)-\mathbf{\bar{A}}\succeq 0$, a vanishing duality gap yields $(\Diag(\bm{y}_\star)-\mathbf{\bar{A}})\mathbf{B}_\star=0$. In particular, the diagonal of this matrix has to vanish: $\Diag(\bm{y}_\star)\Diag(\bm{d}) - \ddiag(\mathbf{\bar{A}}\mathbf{B}_\star) =0$. As it is explained in \cite{Bandeira2016} where an analogous calculation is done, the only solution is $\bm{y}_\star = \Diag(\bm{d})^{-1}\Diag(\mathbf{\bar{A}}\mathbf{B}).$ This yields a dual certificate given by
\begin{equation}
\mathbf{L}(\mathbf{B}) = \Diag(\bm{d})^{-1}\ddiag(\mathbf{\bar{A}}\mathbf{B})-\mathbf{\bar{A}},\label{eq:DualCertificate}
\end{equation}
which certifies the optimality of $\bm{B}$ the following conditions hold $\mathbf{L}(\mathbf{B})\mathbf{B} = 0$ (complementary slackness) and  $\mathbf{L}(\mathbf{B})\succeq 0$ (dual feasibility).


\section{Numerical method}
We address here the numerical solution of the optimization problem \eqref{eq:SDP} thanks to a low rank factorization~\cite{Burer2003}.
Although interior point algorithms can solve SDPs with good theoretical guarantees, we propose here another method with the advantage that it can empirically solve problems of larger sizes. We first use the change of variables  
$\bar{\mathbf{A}}_d= \Diag(\bm{d})^{1/2}\bar{\mathbf{A}}\Diag(\bm{d})^{1/2}$ and $\mathbf{B}_d= \Diag(\bm{d})^{-1/2}\mathbf{B}\Diag(\bm{d})^{-1/2}$.
Then, we factorize $\mathbf{B}_d = \mathbf{H}\mathbf{H^\top}$ with $\mathbf{H}\in\mathbb{R}^{n\times r_0}$ and propose to solve instead:
\begin{equation}
\max_{\mathbf{H}\in \mathbb{R}^{n\times r_0}}\Tr\left(\mathbf{H}^\top\bar{\mathbf{A}}_d\mathbf{H}\right)\text{ subject to }  \|\mathbf{H}_{i\ast}\|_2 = 1, \text{ for all } i\in [n],\label{eq:factorizedProblem}
\end{equation}
where $\mathbf{H}_{i\ast}$ denotes the $i$-th row of $\mathbf{H}\in\mathbb{R}^{n\times r_0}$.  
Inspired by the projected gradient method, a natural projection operator on the feasible $\mathcal{M} = (\mathbb{S}^{r_0-1})^n$ is simply obtained by projecting each factor of the Cartesian product on the unit sphere $ \mathbb{S}^{r_0-1}$, i.e., let $i \in [n]$, then
$[\mathcal{P}(\mathbf{H})]_{i\ast} = \mathbf{H}_{i\ast}/\|\mathbf{H}_{i\ast}\|_2$
normalizes the rows of the matrix $\mathbf{H}\in \mathbb{R}^{n\times r_0}$.  If one row of the matrix $\mathbf{H}$ is a row of zeros, $\mathcal{P}$ returns a random row vector.
Hence, the method in order to maximize $\Tr\left(\mathbf{H}^\top\bar{\mathbf{A}}_d\mathbf{H}\right)$ summarized in Algorithm~\ref{Alg1} consists of a succession of matrix multiplications by $\mathbf{J} = \bar{\mathbf{A}}_d$ and projection steps $\mathcal{P}$.
\begin{algorithm}[h]
\caption{Projected power method~\cite{BoumalGeNPowMeth} \label{Alg1}}
\begin{algorithmic}[1]
\STATE{Input: Symmetric \emph{psd} $\mathbf{J}\in \mathbb{R}^{n\times n}$; and $\mathbf{H_0}\in \mathbb{R}^{n\times r_0}$ such that $\mathcal{P}(\mathbf{H_0}) = \mathbf{H_0}$.}
\FOR{$n =1, 2, \dots$} 
\STATE{$\mathbf{H_n} = \mathcal{P}(\mathbf{JH_{n-1}})$.}
\ENDFOR
\end{algorithmic}
\end{algorithm}
The sequence of iterates of Algorithm~\ref{Alg1} have increasing objective values, as explained in~\cite{BoumalGeNPowMeth}. Once a solution $\mathbf{H}_\star$ is found by using Algorithm~\ref{Alg1}, the embedding coordinates are found by computing the singular value decomposition of $\mathbf{H_d}$. To summarize, the numerical algorithm used to solve a rank constrained version of \eqref{eq:SDP}  is the following:
The initial point for Algorithm~\ref{Alg1} is obtained as $\mathbf{H_0} = \mathcal{P}(\mathbf{M_0})$ where $\mathbf{M_0}\in \mathbb{R}^{n\times r_0}$ is generated with independent entries in $[-1,1]$ chosen uniformly at random. 
Algorithm~\ref{Alg1} yields $\mathbf{H}_\star \in \mathbb{R}^{n\times r_0}$ after convergence and the optimality of the candidate solution $\mathbf{B}_\star = \mathbf{H_\Xi} \mathbf{H_\Xi}^\top$ with $\mathbf{H_\Xi} = \Diag(\bm{d})^{1/2}\mathbf{H_\star}$ can be certified by the dual certificate \eqref{eq:DualCertificate}.
 Finally, a singular value decomposition of $\mathbf{H_\Xi}$ is performed in order to obtain the embedding coordinates.

\section*{Acknowledgments}
M.F. acknowledges stimulating discussions with A. Themelis and thanks the reviewers and associate editor for their suggestions. The authors thank the following organizations. EU: The research leading to these results has received funding from the European Research Council under the European Union’s Seventh Framework Programme (FP7/2007-2013) / ERC AdG A-DATADRIVE-B (290923) and ERC AdG E-DUALITY (787960). This paper reflects only the authors’ views, the Union is not liable for any use that may be made of the contained information. Research Council KUL: Optimization frameworks for deep kernel machines C14/18/068 Flemish Government: FWO: projects:GOA4917N (Deep Restricted Kernel Machines: Methods and Foundations), PhD/Postdoc grant ImpulsfondsAI: VR 2019 2203 DOC.0318/1QUATER KenniscentrumData en Maatschappij Ford KU Leuven Research AllianceProject KUL0076 (Stability analysis and performance improvement of deep reinforcement learning algorithms).

\bibliographystyle{unsrt}
\bibliography{References}

\end{document}